\documentclass[lettersize,journal]{IEEEtran}
\usepackage{amsmath,amsfonts,amssymb}
\usepackage{algorithmic}
\usepackage{algorithm}
\usepackage{array}
\usepackage[caption=false,font=normalsize,labelfont=sf,textfont=sf]{subfig}
\usepackage{textcomp}
\usepackage{stfloats}
\usepackage{url}
\usepackage{verbatim}
\usepackage{graphicx}
\usepackage[hidelinks]{hyperref}
\usepackage{xspace,booktabs}
\usepackage[capitalise]{cleveref}
\usepackage[numbers]{natbib}

\makeatletter
\def\x{{\mathbf x}}
\def\istar{{\mathbf i}^*}
\def\z{{\mathbf z}}
\def\tz{{\widetilde z}}

\def\c{{\mathbf c}}
\def\v{{\mathbf v}}

\def\R{{\mathbb R}}
\def\N{{\mathbb N}}
\def\genc{{\mathord{g}_{\mathrm{enc}}}}
\def\gar{{\mathord{g}_{\mathrm{ar}}}}
\def\gpred{{\mathord{g}_{\mathrm{pred}}}}
\def\gpreds{{\mathord{g}_{\mathrm{pred}}^{(s)}}}

\DeclareRobustCommand\onedot{\futurelet\@let@token\@onedot}
\def\@onedot{\ifx\@let@token.\else.\null\fi\xspace}

\def\eg{\emph{e.g}\onedot} 
\def\ie{\emph{i.e}\onedot} 
 
\def\etc{\emph{etc}\onedot} \def\vs{\emph{vs}\onedot}

\def\ie{\emph{i.e}\onedot}

\def\wrt{w.r.t\onedot}
\makeatother

\title{Bigger is not Always Better: The Effect of Context Size on Speech Pre-Training}
\author{
    Sean Robertson, Ewan Dunbar%
    \thanks{%
        S. Robertson and E. Dunbar are with the University of Toronto.
    
        This work has been submitted to the IEEE for possible publication. Copyright may be transferred without notice, after which this version may no longer be accessible.
    }
}

\begin{document}
%
\maketitle
\begin{abstract}
It has been generally assumed in the automatic speech recognition (ASR) literature that it
is better for models to have access to wider context windows. Yet, many of the potential
reasons this might be true in the supervised setting do not necessarily transfer over
to the case of unsupervised learning.
We investigate how much context is necessary to achieve high-quality pre-trained acoustic models
using self-supervised learning. We principally investigate contrastive predictive coding (CPC),
which we adapt to be able to precisely control the amount of context visible to the model
during training
and inference. We find that phone discriminability in the resulting model representations peaks
at around 40~ms of preceding context, and that having too much context (beyond around 320~ms)
substantially degrades the quality of the representations. Surprisingly, we find that this pattern
also transfers to supervised ASR when the pre-trained representations are used as
frozen input features. Our results point to potential changes in the design of current upstream architectures to better facilitate a variety of downstream tasks.
\end{abstract}


\section{Introduction}
\label{sec:intro}

\IEEEPARstart{A}{ll} machine learning, including that applied to speech processing, involves predicting the values of unknown variables from known ones. In
recent years,
the field has generally assumed
that the more known values that can be fed into the model,
the better.
With the rise in popularity of BERT \citep{devlinBERTPretrainingDeep2019}, many speech systems
have been
built upon the self-attention layer \citep{radfordRobustSpeechRecognition2022,gulatiConformerConvolutionaugmentedTransformer2020,hsuHuBERTSelfsupervisedSpeech2021,baevskiWav2vecFrameworkSelfsupervised2020,baevskiData2vecGeneralFramework2022,chenWavLMLargescaleSelfsupervised2022,zhangGoogleUSMScaling2023}, capable of incorporating
input context at arbitrary distances
into their per-frame outputs. For supervised tasks such as
automatic speech recognition (ASR),
comparisons against models with restricted context have mostly favoured the unrestricted variety \citep{dongSelfattentionAlignerLatencycontrol2019,zhangTransformerTransducerStreamable2020,fuUFO2UnifiedPretraining2023,swietojanskiVariableAttentionMasking2023}. Even if most of the information sufficient in deciding which character or word to transcribe sits within less than a second of its
uttering \citep{assmannVowelIdentificationOrthographic1982,toscanoWithinconsonantErrorsIsolated2014},
additional information (\eg syntactic, semantic, speaker, \etc) which could help in that decision
may sit
outside of that interval; a powerful, supervised speech system can learn what to keep and what to filter out.

Yet while unconstrained context has been largely beneficial to supervised tasks, that boon may not transfer so nicely to
unsupervised speech pre-training.
The goal of a pre-training task is far more nebulous than its supervised counterparts: to train an ``upstream'' model which emits representations (feature vectors) of speech which facilitate
downstream tasks
\citep{mohamedSelfsupervisedSpeechRepresentation2022,borgholtBriefOverviewUnsupervised2022,yangSUPERBSpeechProcessing2021}.
Self-supervision---the
prediction of parts of an input signal from other
parts---has become a commonplace strategy in recent years.
Yet,
it is not at all obvious why the patterns most important to
self-supervised prediction
will extend to
other settings.
In particular, the
calculus of context changes: information relevant to a predictive task could be detrimental to a given downstream task. Despite this possibility, state-of-the-art pre-trained systems continue to use unconstrained self-attention uncritically \citep{gulatiConformerConvolutionaugmentedTransformer2020,hsuHuBERTSelfsupervisedSpeech2021,baevskiWav2vecFrameworkSelfsupervised2020,baevskiData2vecGeneralFramework2022,chenWavLMLargescaleSelfsupervised2022}.

The goal of this research is to examine these one-size-fits-all assumptions of prediction and context in more depth.
%
We orient our experiments around the ABX phone discriminability task \citep{schatzEvaluatingSpeechFeatures2013,hallapEvaluatingContextinvarianceUnsupervised2023}.
The ABX task measures the degree to which an upstream model discriminates between
phonemes, according to the gold labels,
by scoring the number of times
model representations of tokens of a given phoneme
are more similar to other representations
of the same versus another phoneme.
In ideal conditions,
humans take
only a fraction of a second of speech to identify
phonemes
with high accuracy \citep{assmannVowelIdentificationOrthographic1982,toscanoWithinconsonantErrorsIsolated2014}.
While longer time spans encode contextual information which may help the human listener \citep{boothroydMathematicalTreatmentContext1988}, the pre-training objective is not guaranteed to synthesize that context in a way amenable to the task.
%
We hypothesize that
    providing
    too much context as input to
    a self-supervised pre-trained speech model
    is detrimental to
    ABX phoneme discriminability.

The primary goal of this research is to experimentally validate
this hypothesis
and to determine how much input is ``too much.'' To wit, we first construct a
main experimental design wherein
a model and training regime from \citet{nguyenZeroResourceSpeech2020}
is adapted {via} a causal, chunked, self-attention mechanism, to process input at one of seven, fixed-width intervals. We followed \citet{nguyenZeroResourceSpeech2020} for convenience and efficiency: their upstream models, trained on a contrastive predictive coding (CPC) objective \citep{vandenoordRepresentationLearningContrastive2018}, achieved strong ABX performance at relatively small sizes.
We
rigorously evaluate whether our hypothesis holds for these settings, then extend our findings to other settings, including different models, data, and pre-training objectives. In addition to the
ABX phoneme discrimination
task, we
explore
the
effect of
pre-training context width
on
automatic speech recognition
(ASR) performance.


In \cref{sec:related}, we discuss the role restricted context has taken in speech systems thus far, most recently in the form of ``chunking.'' In \cref{sec:methods}, we detail the 
the ABX
task,
the CPC 
pre-training objective 
\citep{vandenoordRepresentationLearningContrastive2018},
and how the chunking mechanism may be leveraged to limit context in self-attention. In \cref{sec:experiments} we cover both the main and auxiliary experimental designs, reporting their results in \cref{sec:results}. We discuss the ramifications in \cref{sec:discussion} and conclude in \cref{sec:conclusions}.

All of our code and 
experimental results
are available online and freely at \url{https://github.com/sdrobert/scpc}.\footnote{%
    Last accessed Decenber 3rd, 2023.%
}





\section{Related Work}
\label{sec:related}

\noindent Restricting processing to a window of audio has been a
mainstay in speech technologies. The filter bank representation, a common time-frequency 
feature, is a vector computed once every (normally) $10$ milliseconds and bound to a region of analysis of $25$ milliseconds. Note that we will refer to any vector extracted from the audio with a fixed period (for our purposes, always $10$~ms) as a ``frame.''
In addition to the windowing done in the features, ASR systems based on
Gaussian mixture models (GMMs)  compute tri-phone distributions with $1$ to $9$ frames' context.
Time delay neural networks (TDNNs), in spite of their rich representational capacity, often end up
using only
a few dozen frames of context (\eg \citep{peddintiJHUASpIRESystem2015}).
Convolutional neural networks (CNNs)
perform local computations with respect to time and frequency, often with much shorter windows in time than in frequency in the case of ASR \citep{abdel-hamidConvolutionalNeuralNetworks2014}. 
In the domain of unsupervised learning, the original wav2vec is a stack of convolutions whose frames are derived from less than a second's context \citep{schneiderWav2vecUnsupervisedPretraining2019}.

In an encoder-decoder architecture, an encoder processes an input sequence (in ASR, speech features) and the decoder ingests the encoder's representations, spitting out its own sequence of some arbitrary length (in ASR, the transcript).
Attention, discussed in \cref{sec:csa},
allows
the decoder to access any part of the input sequence at once, removing the need
to keep track of the entire encoder sequence at once \citep{bahdanauNeuralMachineTranslation2015}. This  ability to handle arbitrarily long-term dependencies made attention an attractive mechanism in general, prompting an explosion of usage in encoder-only
models---notably BERT \citep{devlinBERTPretrainingDeep2019}---which output
sequences of the same length as their input.\footnote{Or something proportional: that is, down-sampling.} Attention found wide use in both ASR and pre-trained speech models (\eg \citep{radfordRobustSpeechRecognition2022,zhangGoogleUSMScaling2023,gulatiConformerConvolutionaugmentedTransformer2020,chenWavLMLargescaleSelfsupervised2022,baevskiData2vecGeneralFramework2022,baevskiWav2vecFrameworkSelfsupervised2020}).

Though the original goal was to model extremely long-term dependencies, researchers have proposed a variety of alterations to attention which limit its extent. Many may be classified as
some form of ``monotonic attention'' which
ensures
that the index of the encoder representation a decoder attends to (\wrt time) increases as the decoder outputs successive tokens. Early methods involve deciding
(somehow)
where to end the attention window per decoder step, attending to the encoder representation between that end and the previous end  \citep{raffelOnlineLineartimeAttention2017,tjandraLocalMonotonicAttention2017}. Segments may be smoothed by parametrizing Gaussian windows \citep{houGaussianPredictionBased2017}, limited to a fixed chunk of past context \citep{chiuMonotonicChunkwiseAttention2018}, or extended backwards indefinitely \citep{arivazhaganMonotonicInfiniteLookback2019}; see \citet{prabhavalkarComparisonSequencetosequenceModels2017} for an empirical comparison of the varieties relevant for ASR. Though monotonic attention is a principled means of limiting context to a few frames, it requires a decoder to govern the windowing
mechanism,
something which pre-trained models do not have.

In encoders for ASR and speech pre-training, a widespread and simple means of limiting attention context is through
chunking:
limiting the input of one or more self-attention layers to a fixed-width, contiguous span of vectors
(a ``chunk''),
then stitching the results of each chunk together during inference. \citet{dongSelfattentionAlignerLatencycontrol2019}
chunked
input in
a
self-attention-based speech recognizer into blocks of 300 to 2000~ms with varying degrees of overlap. Though restricting context to only about $2$ seconds did increase the model's 
character error rate
(CER), it did so by only around $2\%$ relative. \citet{zhangTransformerTransducerStreamable2020}, similar to our approach, used masking to limit attention windows for the purpose of self-attention-only ASR. Though their experiments focus on limiting forward attention to achieve state-of-the-art results, their results when limiting backward attention to only a $5$- to $6$-second window were competitive. \citet{fuUFO2UnifiedPretraining2023} mixed full attention with chunked, causal attention in a joint loss function to pre-train speech representations, yielding comparable results to full attention in a downstream ASR task. \citet{swietojanskiVariableAttentionMasking2023} recently combined backward attention with multi-pass chunking to yield competitive ASR results. Though their best results were obtained with arbitrarily-long context, they achieved an astounding $4.22\%$ WER on in-house data using only about $800$~ms of context.
Finally, Whisper
stitches together transcriptions from $30$-second chunks \citep{radfordRobustSpeechRecognition2022}.

In the above, chunking (and monotonic attention) is almost always justified in terms of latency or tractability. Vanilla attention is computed over the entire input sequence, making it unsuited to situations where low-latency transcriptions are necessary. Further, since attention has quadratic run-time complexity, chunking becomes necessary for very long utterances. As such, chunking is often framed as a trade-off between accuracy and efficiency.

Chunking is not always less accurate than vanilla attention, however. The developers of Google USM illustrate that using only 30-second chunks in training and inference leads to poor ASR results when utterances are a few minutes long \citep{zhangGoogleUSMScaling2023}. These findings are consistent with previous findings in encoder-decoder-based supervised ASR \citep{chorowskiAttentionbasedModelsSpeech2015,chiuComparisonEndtoendModels2019}. It is difficult to assess the severity of the issue empirically, as LibriSpeech \citep{panayotovLibrispeechASRCorpus2015}, the current \emph{de facto} standard contemporary ASR benchmark, is limited to utterances of at most 30 seconds.

Regardless of the reason, to the best of our knowledge, chunking has only been systematically explored in the case of supervised ASR.\footnote{%
    \citet{schneiderWav2vecUnsupervisedPretraining2019} do compare the effects of cropping utterances to variable widths during pre-training on downstream phone recognition performance but, since their models are of a fixed context width already much shorter than those cropped spans, the changes are more likely to be a function of the network's contrastive loss.%
} When chunking has been employed in speech pre-training, results across chunk sizes have not been reported.
One exception is the comparison between $30$-second \vs $8$-second windows in \citep{zhangGoogleUSMScaling2023}.
However, even this comparison is limited, as it evaluates only with respect to downstream ASR performance.
Research which evaluates the role of context windowing on pre-training
in depth
has been absent from speech processing research.

The closest sibling that we have found to our work is that of \citet{tianWhatMakesGood2020}, who argue against the notion of a universal viewpoint in contrastive learning. A ``viewpoint'' refers to a transformation of the input which an upstream model is supposed to learn to ignore as noise. Noise to some downstream task may be valuable information to another (or \emph{vice versa}), hence the lack of universality. As ``chunking'' may be considered a specific instance of transformation, the same argument applies here. The difference between this work and that of \citet{tianWhatMakesGood2020} is primarily the choice of applied domain: the authors illustrate the principle solely on computer vision tasks. That said, we also believe their argument against a universal viewpoint should hold for any pre-training objective, not just those which are contrastive. We provide preliminary support for this in our experimentation.

\section{Methods}
\label{sec:methods}

\noindent We outlined our goals for this research in \cref{sec:intro} and situated them in the current research environment in \cref{sec:related}. As we are looking to investigate whether restricting context in upstream speech models can be beneficial to downstream
evaluations (in our case, phone discriminability),
we cover here the main
methods
we use to do so. We introduce our context-limiting layer in \cref{sec:csa}; the ABX-LS
probe task
in \cref{sec:abx}; and the CPC pre-training objective in \cref{sec:cpc}.

\subsection{Causal, Chunked Self-Attention} \label{sec:csa}

\noindent Attention \citep{bahdanauNeuralMachineTranslation2015,vaswaniAttentionAllYou2017}, in full generality, takes as input a query vector $q \in \R^{d_Q}$, some number $T \in \N$ of key vectors ${\bf k} = (k_1, \ldots, k_T)$, $k_t \in \R^{d_K}$, and the same number of value vectors ${\bf v} = (v_1, \ldots, v_T)$, $v_t \in \R^{d_V}$. Parametrized by a score function $\mathord{score} : \R^{d_Q} \times \R^{d_K} \to \R$, attention produces a vector in $\R^{d_V}$ defined as:
\begin{gather}
    \mathord{attend}(q, {\bf k}, {\bf v}) = \sum_{t=1}^T \alpha(q, {\bf k}) v_t,\mathrm{\>where} \label{eq:att} \\
    \alpha(q, {\bf k}) = \frac{\exp\left(\mathord{score}(q, k_t)\right)}
                              {\sum_{t'=1}^T exp\left(\mathord{score}(q, k_{t'})\right)}.
\end{gather}

There are a variety of score functions to choose from. \citet{vaswaniAttentionAllYou2017} defined the multi-headed, scaled, dot-product attention which we use in all of our experiments.

Self-attention \citep{vaswaniAttentionAllYou2017} merely identifies all input vectors in \cref{eq:sa}
with
elements of a single sequence $\x = (x_1,\ldots,x_T)$, including the query. We can choose the query $T$ different ways, leading to $T$ different output vectors. Thus, self-attention converts an input sequence $\x^{(in)}$ into an output sequence of the same dimensions $\x^{(out)}$:
\begin{equation}
    x_t^{(out)} = \mathord{attend}\left(x_t^{(in)}, \x^{(in)}, \x^{(in)}\right). \label{eq:sa}
\end{equation}

Any output vector $x_t^{(out)}$ depends non-linearly on the entire input sequence $\x^{(in)}$. Since the score function is point-wise, $T$ may change dynamically with new input. Thus, self-attention can be computed on sequences of arbitrary length.

Chunking the self-attention mechanism to limit its dependencies is straightforward: we merely limit the second and third argument of $\mathord{attend}$ to the same, contiguous span of vectors of $\x^{(in)}$. In addition, the CPC loss function (\cref{sec:cpc}) demands the layer be causal: the vector $\x_t^{(out)}$ can only depend on the input vectors at or before the index $t$. Together, our causal, chunked self-attention layer is defined as:
\begin{equation}
    x_t^{(out)} = \mathord{attend}\left(x_t^{(in)}, \x_{\max(t-W,1):t}^{(in)}, \x_{\max(t-W,1):t}^{(in)}\right), \label{eq:csa}
\end{equation}
where $W \in \N$ is the context width and $\x_{a:b} = (x_a, \ldots, x_b)$.

The main mechanism by which we will control the amount of context available to our pre-trained front-ends is the hyperparameter $W$: a representation (an output vector) generated by a single-layer causal, chunked self-attention network depends only on the input vectors at and $W - 1$ frames prior to the index of the representation. Setting $W = T$ in \cref{eq:csa} recovers
an un-chunked causal self-attention layer.

\subsection{ABX-LS Phone Discrimination Benchmark} \label{sec:abx}

\noindent The 
spoken word is comprised of a series of 
phonemes, units which (up to the limits of the combinatorial restrictions of the language), when exchanged one for another, make listeners perceive a different word or possible word.
Critically for our purposes, there is no temporally narrower unit of linguistic contrast.
Hence, phonemes are the ideal subject of a task intended to probe the notion of sufficient, minimal
context. Note that we are talking about \emph{linguistic} contrast.
    Something smaller could be relevant to voice or audio representations, but 
our attention here is on processing relevant to language understanding.

Phonemic discrimination is the basis of the acoustic or sub-word unit task
of the Zero Resource Speech Challenge
\citep{versteeghZeroResourceSpeech2016,dunbarSelfsupervisedLanguageLearning2022},
measured through
a ``machine ABX'' discrimination task, based on methods in psychophysics
\citep{schatzEvaluatingSpeechFeatures2013,hallapEvaluatingContextinvarianceUnsupervised2023}.
Intuitively, two segments of speech perceived as the same phoneme
are
more perceptually similar
than two segments of speech perceived as different phonemes: the ABX task asserts that this similarity should be readily observed in speech representations.
Each sequence of speech representations can be segmented according to the boundaries of phonemes derived from force-alignment. Those sub-sequences are then partitioned into sets labelled with the same phoneme.
We construct a measure of dissimilarity between two spans: consistent with the tradition in the Zero Resource Speech challenge, we do so using dynamic time warping (DTW) using the angular distance as the underlying  divergence score \citep{nguyenZeroResourceSpeech2020} (other approaches, for example using pooling, are possible).
Then, the ABX error rate is calculated based on the frequency with which pairs of spans labelled as the same phoneme are more dissimilar to one another than spans labelled as different phonemes.

Many machine ABX tasks have been constructed, on a variety of languages and with a variety of different purposes.
We 
opt for the ABX-LS suite of
tests for English
introduced in \citet{hallapEvaluatingContextinvarianceUnsupervised2023}, in turn derived from \citet{nguyenZeroResourceSpeech2020}.
ABX-LS is based on force-aligned evaluation partitions of the LibriSpeech corpus, with the corpus itself extracted by automatically aligning audiobook texts with their recordings \citep{panayotovLibrispeechASRCorpus2015}.
The evaluation suite proposed by \citet{hallapEvaluatingContextinvarianceUnsupervised2023} takes representations of isolated phonemes as input, in contrast to earlier ABX evaluations which include additional context in the input (\emph{triphone ABX}), a setting which we put aside except briefly in \cref{sec:exp_main} to establish continuity with the results from the Zero Resource Speech Challenge.

In addition to
\emph{development} and \emph{test}, the evaluation is divided
into \emph{clean} and \emph{other} conditions, depending on how easy the automatic alignment was. 
Second, as the immediate neighbours of the phoneme (\ie, its context) can drastically change its spectral properties, one may ask whether phonemes may be discriminated in any context (or \emph{without} context) or whether discrimination is only 
viable
\emph{within} the same 
phonemic context (defined as one phoneme before and after).
Finally, representations may discriminate between different phonemes \emph{within} speakers, but not necessarily \emph{across} speakers.
Together, these factors yield sixteen distinct evaluation conditions.
%
%
We
find no reason to prefer one
condition over another for the purposes of our hypothesis.
As such, when we report an ABX-LS error rate, it is usually the arithmetic mean of the
scores in the sixteen conditions.
We validate this assumption by searching for interactions among the context widths $W$ and the 
sixteen ABX-LS conditions
in \cref{sec:results}.


\subsection{CPC and Pre-Training} \label{sec:cpc}

\noindent As mentioned in \cref{sec:intro}, self-supervised pre-training involves predicting part of a signal from other parts of the signal. What part specifically is being predicted and how that prediction is operationalized has been used to taxonomize existing approaches to self-supervision \citep{mohamedSelfsupervisedSpeechRepresentation2022}.  Since the majority of our experiments train using
contrastive predictive coding (CPC) \citep{vandenoordRepresentationLearningContrastive2018},
we will tailor our explanation of a
self-supervised pre-training
loss function to CPC. For a more thorough account of the various self-supervised pre-training strategies, consult \citet{mohamedSelfsupervisedSpeechRepresentation2022}.

To describe the approach, we adopt the nomenclature of \citet{vandenoordRepresentationLearningContrastive2018} when available. A CPC-compatible architecture consists of three networks: two main and one auxiliary. The first, ``encoder'' network $\genc$, receives as input the raw audio sequence $\x = (x_1,\ldots,x_{T_1}), x_t \in \R$, and outputs a shorter, ``latent sequence'' of vectors $\genc(\x) = \z = (z_1, \ldots, z_{T_2})$, $z_t \in \R^{H_1}$, $T_2 \ll T_1$. The second, so-called ``auto-regressive'' network, takes in the latent sequence and outputs a ``context sequence'' of vectors $\gar(\z) = \c = (c_1, \ldots c_{T_2}), c_t \in \R^{H_2}$. Note, importantly, that, although \citet{vandenoordRepresentationLearningContrastive2018} use LSTMs for $\gar$, the network need not actually be auto-regressive.%
The third, ``predictor'' network, $\gpred$, receives the context sequence and outputs $S \geq 1$ ``prediction sequences'' $\gpreds(\c) = \v^{(s)} = (v_1^{(s)}, \ldots v_{T_2}^{(s)}), v_t^{(s)} \in \R^{H_3}$, $s \in 1:S$. The prediction vector $v_t^{(s)}$ will be used to predict a latent vector $s$ steps ahead in time, \ie $z_{t+s}$. In \citet{vandenoordRepresentationLearningContrastive2018}, the predictor network was merely a matrix multiplication $v_t^{(s)} = \mathbf{W}^{(s)}c_t$; \citet{nguyenZeroResourceSpeech2020} later generalized $\gpred$ to a single
transformer
layer \citep{vaswaniAttentionAllYou2017} built upon causal self-attention (see \cref{sec:exp_main}). In fact, there is a lot of architectural flexibility in the definitions of $\genc$, $\gar$, and $\gpred$, so long as they generate sequences matching the signatures above. In practice, we will also restrict ourselves to only causal networks as the CPC loss function predicts future values from past ones. Other loss functions have no such restriction and, indeed, it would be an easy thing to modify CPC to remove it. We stick with the standard definition so as to maintain a through-line with previous work.

The goal of CPC pre-training is to transform the audio sequence $\x$ into context vectors $\c$ which, by virtue of the learned transformation $g = \gar \circ \genc$, are
more useful for
downstream
speech tasks.
The prediction network $\gpred$ and its predictions $\v$ serve only as a vehicle for pre-training and are discarded upon its completion.
%
The
intermediate, latent sequence $\z$ 
plays no special role
during inference, but
plays an integral role in the pre-training objective: functionally, $\z$ is the thing being predicted by the prediction network. The objective of self-supervised pre-training is to maximize the similarity between $\v$ and
$\z$ in some sense. 
In the case of CPC, that sense is
``contrastive'':
the degree to which $v_t$ and $z_{t'}$ are similar depends on how strongly $v_t$ and $z_{t'}$ coincide \vs how strongly $v_t$ coincides with the average latent vector. Assuming the latent vectors $z_t$ are of the same length as the prediction vectors $v^{(s)}_t$ and denoting their dot product as $z_t^\intercal v_t^{(s)}$, we have:
%
\begin{gather}
    \mathcal{L}_{\text{CPC}} = \frac{1}{S}\sum_{s=1}^S \mathcal{L}^{(s)}_{\text{CPC}},\text{\>where} \label{eq:cpc} \\
    \mathcal{L}^{(s)}_{\text{CPC}} = - \frac{1}{T-S}\sum_{t=1}^{T - S} \log \frac{\exp\left(z_{t+s}^\intercal v^{(s)}_t\right)}
                                                           {\sum_{\tz} \exp\left(\tz^\intercal v^{(s)}_t\right)}.
                                                                                   \label{eq:cpc_s}
\end{gather}
The sum in the denominator of \cref{eq:cpc_s} is over $M$ latent vectors $\tz$. One of those $M$ vectors is $z_{t+s}$; the remaining $M - 1$ are drawn from a proposal distribution approximating a global prior over possible $z_t$. The proposal distribution is usually taken as the uniform distribution over all $z_t$ generated in the mini-batch. Intuitively, at a given point in time $t$, the prediction network matches some unique part of the latent representation $s$ frames ahead in time, \ie $z_{t+s}$, without matching parts common to the average latent representation. Observing \cref{eq:cpc}, the reader may convince herself that the CPC loss is minimized when the sequential prediction vectors match orthogonal components of $\z$.

\Cref{eq:cpc} provides the definition of the CPC loss function which we will use for the majority of the experimentation covered in \cref{sec:experiments}. Of course, other objectives exist, one of which we will experiment with in that section. Yet despite the surface differences between the objective in \cref{eq:cpc} and the myriad of alternatives 
(see \citep{mohamedSelfsupervisedSpeechRepresentation2022}), 
something akin to the networks $\genc$, $\gar$, and $\gpred$ and the sequences $\x$, $\z$, $\c$, and $\v$, exist across all
methods for doing predictive or contrastive self-supervised learning for speech:
an
input $\x$ 
is 
transformed into
a
sequence
$\z$,
to be predicted. To facilitate this, $\z$ is further transformed into
two other sequences,
$\c$ and 
$\v$, with the only functional difference between them being that $\c$ is kept as input to downstream tasks while $\v$ is discarded.

While we highlight this commonality in part to convince the reader of the similarity between self-supervised pre-training approaches, we do so in other part to emphasize the ambiguity in the distinguished parts: it is not clear why $\z$ ought be distinguished from $\c$, nor $\c$ from $\v$, since $\genc$, $\gar$, and $\gpred$ are, from an architectural standpoint, interchangeable. We believe that the primary purpose of the distinction is to provide degrees of freedom which may be relevant to the quality of the representation, such as the amount of input audio available to a representation $\c$, not explicitly encoded into the pre-training objective. We discuss more in \cref{sec:discussion}. For now, we continue by introducing our experiments and providing their results.

\section{Experiments}
\label{sec:experiments}

\noindent A comprehensive assessment of
our hypothesis
would involve training and evaluating a variety of neural architectures, loss functions, and data over many repeated trials. For feasibility's sake, we decided instead to restrict our attention to three suites of experiments. In the first
suite---the
so-called ``main''
experiments---we
manipulate the main experimental
variable---context width, $W$---of
a CPC model equipped with causal, chunked attention, evaluating the models on the ABX-LS task. 
The second suite, outlined in \cref{sec:exp_aux}, consists of a battery of auxiliary experiments wherein confounding variables in the main setup---namely, the number of layers, the type of layer, and the type of pre-training loss---are manipulated independently. Only one model is trained per context width $W$, in each of these conditions. The goal here is to illustrate that the trends established in the main setup
generalize to other setups. In our final suite of experiments, we take the best models from the main setup (\ie those with the lowest ABX-LS error rate) and use them as pre-trained front-ends for a downstream ASR task. The ASR setup is described in \cref{sec:exp_asr}.


\subsection{Main Setup} \label{sec:exp_main}

\noindent In this section, we introduce the training and architectural decisions most relevant to the design and implementation of our experiments. There are a number of minute details which we have foregone. For a full account, consult our code base.

We designed our experiments by iterating off of the CPC-Small network developed for ZeroSpeech 2021 \citep{nguyenZeroResourceSpeech2020} and its corresponding code base.\footnote{%
    \url{https://github.com/facebookresearch/CPC_audio}, last accessed October 17th, 2023.
} In the original CPC-Small, the encoder network $\genc$ consisted of a stack of $5$ 1D-convolutional layers with kernel sizes $(10,8,4,4,4)$, strides $(5,4,2,2,2)$, and $H_1 = 256$ channels each. The receptive field of $\genc$ was $465$ samples and  generated a $256$-dimensional vector $z_t$ every $160$ samples. At a sampling rate of $16$ kHz, $\genc$ the receptive field and frame shift (\ie the time between $z_t$ and $z_{t+1}$) were roughly $30$ milliseconds and $10$ milliseconds, respectively. The auto-regressive network $\gar$ was a two-layer stack of
long short-term memory
networks (LSTMs) \citep{hochreiterLongShorttermMemory1997}, each of hidden state size $H_2 = 256$. $\gpred$ was defined as a single, causal transformer layer for each of the $S$ prediction steps, a transformer layer consisting of an $8$-headed self-attention sub-layer followed by a feed-forward sub-layer. After each sub-layer was a residual connection \citep{heDeepResidualLearning2016} and layer normalization \citep{baLayerNormalization2016}. The feed-forward sub-layer could change the size of intermediate representations, but remained $256$ so that $v_t^{(s)}$ would match the dimension of $z_t$. To make the transformer layer causal, the self-attention layer is swapped with its causal version. During training, audio was chunked into $128$-frame blocks, then shuffled within-speaker and stored in batches of $64$ blocks. $S = 12$ steps in the future were predicted.

To support our later experimental modifications, we make some small changes to the architecture of \citet{nguyenZeroResourceSpeech2020}.
First, $\gpred$ consists of just one transformer layer for all $S$ prediction steps, but its sub-layer outputs a vector of size $S \times H_2 = 3072$. That vector is then split into $S$ parts for each prediction vector $v_t^{(s)}$. These modifications avoid learning a new self-attention layer for each prediction step $s$, drastically speeding up training and reducing the model's GPU memory footprint. Second, rather than shuffling blocks within speaker, the entire corpus is processed sequentially for simplicity and to reduce memory pressure. For comparison, we train this modified model for $200$ epochs on the \emph{train-clean-100} subset of LibriSpeech with a small subset of that partition separated as validation data according to \citet{vandenoordRepresentationLearningContrastive2018}. We keep the checkpoint with the lowest validation loss over all epochs.

\begin{table}
    \caption{%
        Comparing ABX-LS error rates (\%) of implementations of CPC-Small on a subset of the triphone-context conditions. \emph{dev-clean} and \emph{dev-other} are partitions of LibriSpeech; \emph{within} and \emph{across} refer to discriminating pairs of triphones from the same or across speakers. 
    } \label{tab:cpc_small}
    \centering
    \begin{tabular}{r c c c c }
        \toprule
           & \multicolumn{2}{c}{dev-clean} & \multicolumn{2}{c}{dev-other} \\
                                                & within & across & within & across \\
        \midrule
        \citet{nguyenZeroResourceSpeech2020}    & 6.2    & 8.5    & 8.2    & 13.6 \\
        Ours                                    & 6.2    & 8.9    & 8.5    & 14.9 \\
        Ours (no layer norm)                    & 8.7    & 12.0   & 8.7    & 17.0 \\
        \bottomrule
    \end{tabular}
\end{table}

The first two rows of \cref{tab:cpc_small} compare the reported results of the official CPC-Small implementation to those of our implementation. Note that \cref{tab:cpc_small}
lists triphone discrimination error rates rather than those for phonemes as \citet{nguyenZeroResourceSpeech2020} predates the phoneme measure. For comparison with other results in this section, the mean, phoneme-level ABX-LS error rate for our CPC-Small is $9.6$\%. As we can see, our implementation performs slightly worse than the reference implementation with a $0$-$5\%$ relative increase in error rates within speaker and $5$-$10\%$ increase across, but, overall, the results are comparable.

This modified architecture allows us to carry out our experiments as follows. We replace the LSTMs in $\gar$ with a single causal, chunked, transformer layer followed by a final, feed-forward layer. The feed-forward sub-layer in the transformer layer is of hidden state size $1024$, but the final feed-forward layer is of size $256$, returning the speech representations to the same size as those of CPC-Small. The increase in hidden state size is chosen to achieve near parity in the number of parameters (around $2.1$ million) with CPC-Small (around $2.4$ million). Our main experimental variable is the context width $W$ of the causal, chunked, self-attention sub-layer of the transformer layer, which does not change the size of the model.

We experimented with widths $W \in \{2^i|i\in 1:7\}$,\footnote{%
    $W=1$ would lead to an identity mapping in self-attention.%
} corresponding to durations of roughly
40 to 1300~ms.
As the average phone duration in LibriSpeech appears to be around
90~ms (standard deviation 50),\footnote{%
    Based on the force-aligned durations of phonemes from the \emph{train-clean-100} partition of LibriSpeech, provided by \citet{vandenoordRepresentationLearningContrastive2018}.%
} we felt this range ran the gamut between too 
little context, on the one hand,
and more context than
necessary, on the other.

When experimenting with context widths, we found it necessary to modify two ingredients of the training recipe.\footnote{%
    This involved much trial-and-error. In these erroneous trials, however, we saw no trends contradicting those we present here. The reader may check for herself: our code-base stores the artifacts of nearly all of our experiments.%
} First, pre-chunking utterances into blocks of $128$ frames meant our longer context widths $W$ would often be unrealized due to block boundaries. Offsetting the first frame we compute the loss for from the start of the block mitigates this issue, but would expose longer windows to fewer training examples. Instead, we removed chunking altogether, training on $12$ entire utterances per
batch---the
rough equivalent to $64$ blocks.

Second, we found it necessary to disable layer (channel) normalization in-between convolutional layers. The third row of \cref{tab:cpc_small} lists the ABX-LS performance of our implementation of CPC-Small trained without layer normalization. For reference, its mean phoneme-level error rate is $11.6$\%. As we can see, layer normalization leads to much lower error rates, agreeing with the findings of \citet{riviereUnsupervisedPretrainingTransfers2020}. Indeed, we saw much the same in our experiments with context widths. However, layer normalization would also cause networks to plateau in validation loss for an indeterminate number of epochs at the beginning of training. In an extreme example, models being trained on the entire LibriSpeech corpus failed to escape the plateau after weeks of training. As layer normalization often led models to fail to finish training before the $200$-epoch budget, increasing the variance of our results, we decided to disable it. We recommend re-enabling layer normalization in a production environment.

\subsection{Auxiliary Setups} \label{sec:exp_aux}

\noindent After designing the main suite of experiments, we considered the various decisions we had made which, if made differently, could lead plausibly to different outcomes. Such confounding variables fall into three broad classes: those involving the model itself, the pre-training task, and the training regime. Within each of these categories alone is an infeasible number of variables to explore. Thus, we focused on variables which we believed to be the most high-impact while remaining otherwise close to the main setup. We enumerate and justify our choices here.

We begin with the model variables. As modern ML tends towards larger
models---much larger, comparatively, than ours---we
decided to manipulate size first. As the attention layers were already inflated to match the size of CPC-Small, we decided instead to adjust the number of layers to increase our model size. Stacking layers (with suitable point-wise non-linearities in-between) can increase the expressivity \citep{hornikMultilayerFeedforwardNetworks1989} and also the stability \citep{mallatUnderstandingDeepConvolutional2016} of learned models. Stacking $D$ layers, each of width $W$, also increases the total context width contributing to each representation vector to
$D(W - 1) + 1$ frames.
Thus, layers provide an alternative, more expressive means of introducing context to a network. The hyperparameters remain the same as in \cref{sec:exp_main}, except the transformer layer in $\gar$ is replaced with $D \in \{2, 4\}$ such layers, fed one into the other. The $2$-layer network contains around $2.9$ million parameters whereas the $4$-layer network contains around $4.5$ million.

Second, we may manipulate the type of layer. While self-attention is
a popular choice, convolutional layers, as discussed in \cref{sec:related}, still play an integral role in contemporary speech processing. There is also a clear analogous role between the window size $W$ in causal, chunked self-attention and a 1D convolution's kernel width. The difficulty in ensuring comparability comes from the number of parameters: in the convolutional case, a filter contains $H_1 \times W \times H_2$ learnable coefficients, proportional to the context width $W$. To account for this complication, we ran two sets of experiments: one, keeping the total number of parameters fixed while allowing $H_2$ to vary; and the other, keeping $H_2$ fixed while allowing the total number of parameters to grow with $W$. Model sizes ranged between around $1.4$ million parameters ($W = 2$) and $9.7$ million parameters ($W = 128$). For the fixed-parameter experiment, $H_2 = 256$ when $W = 8$. Those models contained around $2.4$ million parameters.\footnote{%
    Despite the name, models in the fixed-parameter condition did vary by a few thousand parameters total. This was due to the convolution's bias vector, which is proportional to $H_2$.%
}

Models incorporating more context have traditionally needed more time and data to converge. As such, we manipulate the length and data available for training. In the former case, we simply extended the duration of training until all models reached a plateau in loss, which occurred within a $500$-epoch budget. In the latter, we train models on the entire $960$-hour LibriSpeech training set. Though models were able to converge within $100$ epochs, the nearly 10-fold increase in training data drastically increased the wall time. As a result, we trained on only two widths: $W \in \{8, 128\}$.

Finally, we consider manipulations to the pre-training objective. First, we consider modifying the CPC objective, which is easily done through the number of prediction steps $S$. While the value $S = 12$ was experimentally derived in \citet{vandenoordRepresentationLearningContrastive2018}, it was done so for CPC-Small (a recurrent model). \citet{shainAcquiringLanguageSpeech2020} performed a similar evaluation, predicting frames in both the backward and forward direction. Their results showed a preference for $S = 10$ forward steps in English phoneme classification, although the optimal value varied by language, task, and architecture. In addition to the $S=12$ prediction steps from the main setup, we trained a model for each $W$ and each number of steps $S \in \{1,3,6,24\}$. We also experiment with an adjustment to the loss function,
\begin{equation}
    \mathcal{L}_{\text{CPC-Last}} = \mathcal{L}^{(S)}_{\mathrm{CPC}}, \label{eq:cpc_last}
\end{equation}
which predicts only the frame $S$ steps away, rather than all frames between $1$ and $S$. \Cref{eq:cpc_last} is not only faster to compute than \cref{eq:cpc}: it also presents a clearer picture of how great a look-ahead is necessary for strong performance on an ABX task.

Second, we consider an alternative to the CPC objective: BEST-RQ \citep{chiuSelfsupervisedLearningRandomprojection2022}. Much like HuBERT \citep{hsuHuBERTSelfsupervisedSpeech2021}, BEST-RQ predicts masked labels from quantized representations of speech frames. Formally, given a discrete code-book of code words $i \in (1,\ldots,H_3)$ and some mapping of the audio $\x$ to the ground-truth sequence $\istar = (i^*_1,\ldots,i^*_{T_2})$, each vector in a single prediction sequence $\v^{(1)}$ minimizes the per-frame categorical pseudo-probability of the sequence $\istar$:
\begin{equation}
   \mathcal{L}_{\text{cat}} = -\frac{1}{T} \sum_{t=1}^{T} \log \frac{\exp(v^{(1)}_{t,i_t^*})}
                                                                       {\sum^{H_3}_{i=1} \exp(v^{(1)}_{t,i})}. \label{eq:lcat}
\end{equation}
If $i$ were to index tokens in a fixed vocabulary of, \eg, words, \cref{eq:lcat} would be nearly identical to the BERT objective \citep{devlinBERTPretrainingDeep2019}. To make the task harder, sub-sequences of latent vectors $\z$ are randomly chosen to be masked out (\eg by setting $z_t$ to all $0$ or Gaussian noise), forcing the model to infer the missing parts of the signal. BEST-RQ was chosen over alternatives for its simplicity: $\z$ is a filter-bank representation of $\x$; $H_3$ prototypical vectors $z^{(i)}$ are drawn per-coordinate from the standard normal distribution $\mathcal{N}(0, 1)$; and the sequence $\istar$ is derived from the indices of the nearest prototype vectors to each (unmasked) $z_t$:\footnote{%
    $z_t$ are first normalized to have unit norm.%
}
\begin{equation}
    i_t^* = \arg\min_i \left\|z_t - z^{(i)}\right\|. \label{eq:bestrq_diff}
\end{equation}

To convert our CPC networks to BEST-RQ networks, we do as follows. We replace the stack of convolutions in $\genc$ with 80-dimensional, log, Mel-scaled feature vectors; we set the code-book size to $H_3 = 8192$; and we mask $S = 12$ frames of contiguous feature vectors with probability $p = 0.01$ per frame. Besides the loss function and the size of the final feed-forward output ($H_3$), $\gar$ and $\genc$ have the same structure as in \cref{sec:exp_main}.\footnote{%
    These systems which learn by masking often allow the self-attention layers in $\genc$ full access to the sequence, besides the masked parts. That is, unlike CPC, their self-attention layers need not be causal. We ultimately decided to stick with causal predictor networks: a) to avoid changing more variables than necessary to evaluate the BEST-RQ loss; and b) because we saw no obvious reason why an unconstrained self-attention layer would benefit one context width over another.%
} At around $0.8$ million parameters, the resulting BEST-RQ networks are around a third of the size of the CPC networks. As such, the networks are not directly comparable; results in \cref{sec:results} should not be construed as empirical support for one method over the other.

\subsection{ASR Task} \label{sec:exp_asr}

\noindent The ABX-LS task is
a probe task which attempts to measure the intrinsic linguistic quality of pre-trained representations, but is only one of many possible evaluations. Even for downstream tasks which in principle rely only on phonemic information, it may give an incomplete picture.
In particular, given that automatic speech recognition (ASR) typically benefits greatly from models that incorporate extensive context (both in end-to-end models and in extrinsic language models),  we might expect that
ASR could benefit from pre-training architectures
that allow for the integration of larger
amounts of context than are necessary for mere phoneme discrimination.

Our ASR task is comparable to that of SUPERB \citep{yangSUPERBSpeechProcessing2021}.
Some pre-trained model generates context vectors $\c$ from each audio sequence $\x$ in LibriSpeech's \emph{train-clean-100} partition. Those context vectors are treated as feature vectors in an otherwise vanilla supervised ASR task. We picked the model with the lowest ABX-LS error rate under the default setup to represent each context width. We optimize a 2-layer, bi-directional, LSTM stack with 1024 hidden units per direction on a
connectionist temporal classification (CTC) loss function \citep{gravesConnectionistTemporalClassification2006}.
We use a sub-word vocabulary of size 2000 generated by applying
byte-pair encoding to the LibriSpeech language model (LM) training data.
We stack 3 context vectors together as input to the LSTM at each step, decreasing the
combinatorial
complexity of the CTC marginalization procedure and therefore speeding up training. To improve model robustness, dropout was applied to hidden states while SpecAugment \citep{parkSpecAugmentSimpleData2019} was applied to the input context vectors. The model was trained until greedy decoding failed to produce a $0.1\%$ drop in sub-word error rate for 10 epochs. Decoding was performed with the Pyctcdecode \citep{kenshotechnologiesPyctcdecode} Python package, which modifies the standard CTC 
prefix search \citep{gravesConnectionistTemporalClassification2006}
to incorporate word-level LMs. We report two sets of results: one set with a $4$-gram LM trained with KenLM \citep{heafieldKenLMFasterSmaller2011} on LibriSpeech LM training
data,
and one set without any LM. LM mixing coefficients and transcript length penalties were tuned to each model's performance on the \emph{dev-clean} partition; the beam width was fixed to $32$.

\section{Results} \label{sec:results}

\begin{figure*}
\centering
\includegraphics[width=\textwidth]{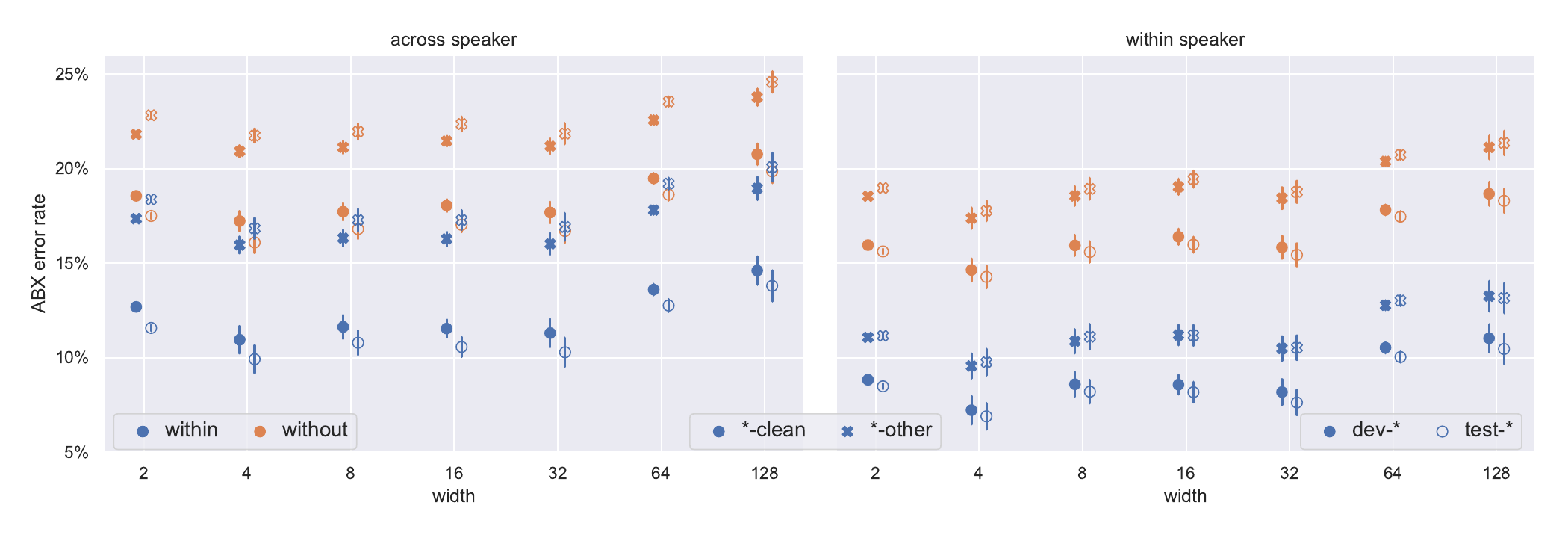}
\caption{%
    Comparing phoneme-level ABX-LS error rates across the various context widths $W$ and varieties of measure. Error bars are $\pm 1$ SE. $N=5$ models were trained per width.%
} \label{fig:abx_ctx}
\end{figure*}

\noindent We begin with the results of the main setup described in \cref{sec:exp_main}. First, however, we address whether it is reasonable to aggregate our phoneme-level ABX-LS scores by looking for interaction effects across the various rate conditions (see \cref{sec:abx}). \Cref{fig:abx_ctx} plots the mean and standard error of error rates by width, speaker condition, partition, and phonemic context. While all factors have a considerable impact on reported scores, none of the factors appear to interact with context width. Thus, taking the mean across ABX-LS error rates is justified when looking for the main effect of context width.

\begin{table*}
    \centering
    \caption{%
        Comparing ABX-LS error rates (\%) across a various model configurations \vs context window lengths ($W$). Error rates are averaged across all phoneme-level measures. For \emph{main} (the main results), $N=5$ runs were averaged. Otherwise, $N=1$.%
    } \label{tab:stats}
    \begin{tabular}{r c c c c c c c}
        \toprule
                            & $2$        & $4$        & $8$        & $16$       & $32$       & $64$       & $128$ \\
         \midrule
         main               & $15.6$     & $\bf 14.2$ & $15.1$     & $15.3$     & $14.8$     & $16.9$     & $17.7$\\
         main (best)        & $15.3$     & $\bf 12.6$ & $13.3$     & $13.8$     & $13.6$     & $16.2$     & $15.9$\\\midrule
         2-layer            & $13.7$     & $15.5$     & $15.8$     & $\bf 12.6$ & $13.6$     & $16.8$     & $13.0$\\
         4-layer            & $13.8$     & $\bf 13.0$ & $14.1$     & $15.8$     & $15.8$     & $16.8$     & $17.1$\\
         conv (fixed param) & $17.0$     & $\bf 14.6$ & $16.8$     & $17.0$     & $19.0$     & $19.4$     & $20.7$\\
         conv (fixed $H_2$) & $15.0$     & $14.6$     & $16.9$     & $\bf 14.2$ & $50.0$     & $22.3$     & $43.7$\\
         long train         & $13.4$     & $\bf 13.0$ & $14.0$     & $14.4$     & $14.3$     & $14.7$     & $15.0$\\
         960h               & $-$        & $-$        & $\bf 17.2$ & $-$        & $-$        & $-$        & $18.5$\\
         last at $S=6$      & $14.0$     & $13.9$     & $\bf 13.1$ & $15.4$     & $13.3$     & $14.9$     & $16.0$\\
         BEST-RQ            & $27.6$     & $\bf 24.1$ & $24.5$     & $26.6$     & $26.1$     & $28.0$     & $25.2$\\
         \bottomrule
    \end{tabular}
\end{table*}

The first row of \cref{tab:stats}---denoted \emph{main}---lists mean  ABX-LS error rates across the various context widths. Among those $N=5$ models, the best (\ie lowest average error rate) is reported in the next row, \emph{main (best)}. The remaining rows correspond to the auxiliary setups from \cref{sec:exp_aux} which we will discuss after considering the main results. The main rows suggests an optimal performance at $W = 4$, comparable performances for $2 \leq W \leq 32$, and a marked decrease in performance as $W$ grows to $64$ frames.

We note that the average performance of even our best model ($12.6\%$) is considerably worse than our implementation of CPC-Small ($9.6\%$), though, without the layer normalization, the results were much closer ($11.6\%$). These results are promising, given that we tuned neither the architecture nor training regime to the experimental models.


Experimental statistics were computed with the Python package Pingouin \citep{vallatPingouinStatisticsPython2018} on the average error rate of the $N=25$ models. A one-way analysis of variance (ANOVA) on the $N=25$ runs showed a significant effect of the choice of context width on ABX-LS error rates, $F(6, 28) = 6.026, p < 0.001$.\footnote{%
    Tests of non-normality and heteroscedasticity did not reach significance.%
}

Wilcoxon signed-rank tests between all pairs of context widths were employed as a \emph{post-hoc} means of probing the source of the significant effect. The only comparisons to reach $p < 0.05$ were those between a shorter window $W \leq 32$ and a longer window $W \geq 64$. Adducing our findings from \cref{tab:stats}, we infer that there is a large, reliable schism in the performance of the systems with context widths lying somewhere between $32$ and $64$ frames, or $0.34$-$0.66$ seconds, with performance favouring the shorter widths. Among the shorter widths (and longer widths), differences in performance are small and variable enough so as not to be reliable. As there is considerable variance in the duration of phonemes in the corpus, there is likely no perfect setting. Nonetheless, windows at or about $64$ frames appear to have an order too much context. We also find a numerical trend suggesting ``too little'' context, with higher error rates at $W = 2$.

We now turn to the results of the auxiliary experiments from \cref{sec:exp_aux}, whose mean error rates are listed in the remaining rows of \cref{tab:stats}. Note that, unlike in the main setup, only one model was trained per cell. Model variants are listed as rows \emph{2-layer}, \emph{4-layer}, \emph{conv (fixed param)}, and \emph{conv (fixed $H_2$)}. \emph{conv (fixed param)} is the convolutional variant where the total number of parameters is kept nearly fixed while \emph{fixed $H_2$} keeps the channels fixed. \emph{Long train} lists results for the increased, $500$-epoch training budget on LibriSpeech's train-clean-100 partition whereas \emph{960h} refers to the results on its full training set. \emph{last at $S = 6$} lists the results of models trained using \cref{eq:cpc_last}, looking ahead $S = 6$ frames. 

Auxiliary results continue to support
our hypothesis,
albeit with much greater variance due to the lack of repeated measures. It is worth noting that $W = 128$ performs admirably in the \emph{2-layer} and \emph{BEST-RQ} rows. We attribute this to luck: $W = 128$ showed the greatest variance in our experiments, which is particularly noticeable in \cref{fig:abx_ctx}. The multi-layer results suggest that stacking layers could be employed to increase total context width\footnote{%
    Recall: $D(W - 1) + 1$.%
} while retaining decent results, though the best-performing models still have total widths less than $64$ frames. Overall, the best-performing system always has a width within $W \in \{4,8,16\}$, showing a continued preference for widths which are neither too short nor long.

\begin{table}
    \centering
    \caption{%
        Comparing ABX-LS error rates (\%) across CPC loss flavours \vs number of prediction steps $S$. \emph{avg}: averaging over prediction steps up to $S$, \cref{eq:cpc}; \emph{last}: only the last step $S$, \cref{eq:cpc_last}.%
    } \label{tab:steps} %
    \begin{tabular}{r c c c c c}
        \toprule
             & $1$    & $3$    & $6$            & $12$       & $24$ \\
        \midrule
        avg  & $43.8$ & $26.1$ & $16.6$         & $\bf 15.7$ & $20.6$ \\
        last & $43.8$ & $17.5$ & $\bf 14.4$     & $16.0$     & $31.6$ \\
        \bottomrule
    \end{tabular}
\end{table}

Though we experimented with all steps $S \in \{1,3,6,12,24\}$  using both \cref{eq:cpc,eq:cpc_last}, only the results for \cref{eq:cpc} on $S = 12$ (\ie \emph{main}) and \cref{eq:cpc_last} are reported in \cref{tab:stats}. As \cref{tab:steps} illustrates, these were the best-performing configurations overall.\footnote{%
    $S \times W$ results are presented in the supplementary materials. We saw no obvious interaction effects. The best results for $W \geq 64$ match the best configurations overall, except the combination $(W, S) = (128, 6)$ on \cref{eq:cpc}. It managed to achieve an average error rate of $17.3\%$ versus the $17.7\%$ from \cref{tab:stats}. The rankings, however, remain the same.%
} We can see that removing averaging drops the optimal choice of $S$ by half, and leads to a much steeper increase in error rates as $S$ increases. Interestingly, the modified CPC objective compares favourably to the original. Given its fewer parameters and cheaper run time, \cref{eq:cpc_last} is a promising alternative to vanilla CPC.



\begin{table*}
\centering
\caption{%
    WERs (\%) of downstream ASR models fed features from upstream CPC models from the \emph{main (best)} configuration. CPC-Small from \cref{sec:exp_main} is included for comparison.
} \label{tab:asr}
\begin{tabular}{r c c c c | c c c c }
    \toprule
                   & \multicolumn{4}{c}{without LM} & \multicolumn{4}{|c}{with LM} \\
                   & dev-clean  & dev-other  & test-clean & test-other & dev-clean  & dev-other  & test-clean & test-other \\
     \midrule
     CPC-Small     & $26.6$     & $50.3$     & $25.5$     & $54.1$     & $14.7$     & $36.4$     & $14.1$    & $40.7$ \\
     \midrule
     $W=2$         & $30.3$     & $51.5$     & $28.9$     & $55.6$     & $19.1$     & $39.4$     & $18.0$     & $43.9$ \\
     $W=4$         & $\bf 26.5$ & $\bf 48.0$ & $\bf 25.1$ & $\bf 51.8$ & $\bf 16.5$ & $\bf 36.6$ & $\bf 15.4$ & $41.0$ \\
     $W=8$         & $29.8$     & $51.1$     & $27.7$     & $54.7$     & $17.0$     & $37.3$     & $16.1$     & $\bf 40.9$ \\
     $W=16$        & $30.9$     & $51.4$     & $29.1$     & $55.2$     & $19.4$     & $39.9$     & $17.9$     & $43.9$ \\
     $W=32$        & $30.8$     & $52.1$     & $29.3$     & $55.8$     & $18.1$     & $38.4$     & $17.1$     & $42.6$ \\
     $W=64$        & $34.3$     & $54.8$     & $33.2$     & $58.7$     & $22.7$     & $42.9$     & $21.3$     & $48.5$ \\
     $W=128$       & $31.9$     & $52.0$     & $30.7$     & $55.5$     & $19.9$     & $39.6$     & $18.9$     & $44.2$ \\
     \bottomrule
\end{tabular}
\end{table*}

We now determine whether our results extend to ASR. \Cref{tab:asr} lists word error rates (WERs) of the models, factored by context widths, partitions, and whether the LM was included in decoding. The pre-trained front-ends correspond to the models from the \emph{main (best)} row from \cref{tab:stats}. In addition, we trained an additional ASR system on the representations from our implementation of the layer-normalized CPC-Small (discussed in \cref{sec:exp_main}). The overall scores outperform SUPERB's CPC baseline ($20.2\%$ on \emph{test-clean}), which is slightly smaller than ours (by around $500$MB) but trained on Libri-Light.  While much larger models (e.g., data2vec \citep{baevskiData2vecGeneralFramework2022}) obtain substantially better scores, the performance is in line with the scope of our experiments, which was limited to variants of CPC. Considering the main manipulation, the results continue to favour shorter windows: $W=4$ beats out even CPC-Small (without an external LM) while $W = 64$ in particular falls behind. Mixing an external LM into the decoding process led to considerably reduced WERs across the board, though CPC-Small overtook $W=4$ in performance. All models performed much better on the \emph{clean} evaluation data than on \emph{other}. Overall, the ranking of models in terms of ABX performance seem to roughly correspond to their ASR performance.

\section{Discussion}
\label{sec:discussion}

\noindent Increased context size is not always beneficial to downstream performance. We saw strong support for the hypothesis
that too much context is detrimental. The two models with the longest context widths ($W=64$ and $W=120$; approximately $640$ms and $1.3$s, respectively) were routinely outperformed by shorter windows, chiefly within the order of a single phoneme (\ie tenths of seconds). The evidence suggests, therefore, that ``too much'' context for phoneme discrimination sits somewhere slightly over the duration of a phoneme. This makes intuitive sense, given the nature of the task.

Our preliminary findings also suggest the ASR downstream task also shows a preference for short windows in pre-training. This might be surprising. ASR, considering the whole pipeline, demonstrably benefits from more context. There are several possible reasons that increasing the context in the pre-training model might not be helpful. First, in this case, the downstream model is very bulky (around $44$ million parameters). Because the downstream model can perform significant pattern recognition on its own, it may be more important for the upstream model to retain a high-fidelity representation of the input than to perform extensive pre-processing. This result is reminiscent of the classic work of \citet{mohamedAcousticModelingUsing2012}, who found that neural speech recognizers trained on spectral features performed better than those trained on cepstral features, the latter involving strictly more pre-processing than the former. Second, our pre-training predicts relatively short windows in the future (up to 240~ms, and here we concentrated on 120~ms). Longer-term context can be useful for resolving ambiguities at the level of whole words or phrases, but windows of this shorter size are not likely to contain information that the more immediate context does not better predict. Finally, perhaps phoneme-sized representations are sufficient context for an ASR encoder or pre-trained acoustic model. If a downstream model can perform predictions based on context, it may be the job of the speech representation merely to faithfully model phonemes.

While our results show a clear preference for speech representations derived from input at roughly the duration of the phoneme, we hesitate to prescribe constraining context in upstream models more generally. There are tasks, such as speaker identification or speech translation, which clearly require more context to achieve.\footnote{%
    Assuming the downstream model is not powerful enough to do the work itself.%
} Moreover, as illustrated by the consistent gaps in performance across ABX-LS conditions in \cref{fig:abx_ctx}, a short window does not guarantee a stable representation across phonemic contexts nor speakers. Thus, even if our results suggest a preference for shorter widths on these tasks using these objectives, we may still do better on other tasks or using other objectives.

It is not obvious how existing predictive pre-training objectives are any more capable of learning a ``universal viewpoint'' \citep{tianWhatMakesGood2020} than CPC or BEST-RQ. Our recommendation, then, is to side-step the problem by learning a heterogeneous representation over multiple viewpoints, allowing downstream models to pick and choose the parts they need. As a starting point, we may train a number of fixed-context upstream models at different widths $W$ and step sizes $S$, and merely concatenate their frame-wise representations. Such an approach could integrate the power and flexibility of self-supervised learning with the classic intuitions and formalism of multi-resolution analysis \citep{mallatWaveletTourSignal2008}. Naturally, we leave any such innovations to future work.




\section{Conclusions}
\label{sec:conclusions}

\noindent While ``more is better'' has been an effective mantra in the realm of supervised learning, it does not transfer so well to self-supervised learning, as existing, predictive, pre-training tasks do not map perfectly to downstream goals. We demonstrated that the amount of input with which speech representations are computed can have a significant effect downstream performance. \emph{Post-hoc} testing was consistent with the hypothesis that there is such a thing as ``too much input,'' with shorter windows preferred over longer ones. That trend was robust to a number of training and architectural variations, including training length, the number of model layers, the type of layer, modifications to the CPC pre-training objective, an entirely new objective, and to the amount of training data. Furthemore, they generalized from phoneme discrimination to the more complex ASR task. These results highlight the need pre-training objectives to better align with the needs of their downstream tasks.

\section{Acknowledgements}
\label{sec:acknowledgements}

\noindent This research is funded by the Data Sciences Institute at the University of Toronto and by the Natural Sciences and Engineering Research Council of Canada (NSERC) RGPIN-2022-04431. It was also enabled in part by support provided by Compute Ontario (\url{https://www.computeontario.ca/}) and the Digital Research Alliance of Canada (\url{https://alliancecan.ca/}). Resources used in preparing this research were provided, in part, by the Province of Ontario, the Government of Canada through CIFAR, and companies sponsoring the Vector Institute (\url{https://vectorinstitute.ai/partnerships/current-partners/}).%

\bibliographystyle{IEEEtranN}
\bibliography{refs}

\begin{thebibliography}{54}
\providecommand{\natexlab}[1]{#1}
\providecommand{\url}[1]{#1}
\csname url@samestyle\endcsname
\providecommand{\newblock}{\relax}
\providecommand{\bibinfo}[2]{#2}
\providecommand{\BIBentrySTDinterwordspacing}{\spaceskip=0pt\relax}
\providecommand{\BIBentryALTinterwordstretchfactor}{4}
\providecommand{\BIBentryALTinterwordspacing}{\spaceskip=\fontdimen2\font plus
\BIBentryALTinterwordstretchfactor\fontdimen3\font minus
  \fontdimen4\font\relax}
\providecommand{\BIBforeignlanguage}[2]{{%
\expandafter\ifx\csname l@#1\endcsname\relax
\typeout{** WARNING: IEEEtranN.bst: No hyphenation pattern has been}%
\typeout{** loaded for the language `#1'. Using the pattern for}%
\typeout{** the default language instead.}%
\else
\language=\csname l@#1\endcsname
\fi
#2}}
\providecommand{\BIBdecl}{\relax}
\BIBdecl

\bibitem[Devlin et~al.(2019)Devlin, Chang, Lee, and
  Toutanova]{devlinBERTPretrainingDeep2019}
J.~Devlin, M.-W. Chang, K.~Lee, and K.~Toutanova, ``{{BERT}}: {{Pre-training}}
  of deep bidirectional transformers for language understanding,'' in
  \emph{Proceedings of the 2019 {{Conference}} of the {{North American
  Chapter}} of the {{Association}} for {{Computational Linguistics}}: {{Human
  Language Technologies}}}, vol.~1, {Minneapolis, USA}, 2019, pp. 4171--4186.

\bibitem[Radford et~al.(2022)Radford, Kim, Xu, Brockman, McLeavey, and
  Sutskever]{radfordRobustSpeechRecognition2022}
A.~Radford, J.~W. Kim, T.~Xu, G.~Brockman, C.~McLeavey, and I.~Sutskever,
  ``Robust speech recognition via large-scale weak supervision,'' 2022.

\bibitem[Gulati et~al.(2020)Gulati, Qin, Chiu, Parmar, Zhang, Yu, Han, Wang,
  Zhang, Wu, and Pang]{gulatiConformerConvolutionaugmentedTransformer2020}
A.~Gulati, J.~Qin, C.-C. Chiu, N.~Parmar, Y.~Zhang, J.~Yu, W.~Han, S.~Wang,
  Z.~Zhang, Y.~Wu, and R.~Pang, ``Conformer: {{Convolution-augmented}}
  transformer for speech recognition,'' in \emph{Proc. {{Interspeech}} 2020},
  2020, pp. 5036--5040.

\bibitem[Hsu et~al.(2021)Hsu, Bolte, Tsai, Lakhotia, Salakhutdinov, and
  Mohamed]{hsuHuBERTSelfsupervisedSpeech2021}
W.-N. Hsu, B.~Bolte, Y.-H.~H. Tsai, K.~Lakhotia, R.~Salakhutdinov, and
  A.~Mohamed, ``{{HuBERT}}: {{Self-supervised}} speech representation learning
  by masked prediction of hidden units,'' \emph{IEEE/ACM Transactions on Audio,
  Speech, and Language Processing}, vol.~29, pp. 3451--3460, 2021.

\bibitem[Baevski et~al.(2020)Baevski, Zhou, Mohamed, and
  Auli]{baevskiWav2vecFrameworkSelfsupervised2020}
A.~Baevski, Y.~Zhou, A.~Mohamed, and M.~Auli, ``wav2vec 2.0: {{A}} framework
  for self-supervised learning of speech representations,'' in \emph{Advances
  in {{Neural Information Processing Systems}} 33}, H.~Larochelle, M.~Ranzato,
  R.~Hadsell, M.~Balcan, and H.~Lin, Eds., vol.~33.\hskip 1em plus 0.5em minus
  0.4em\relax {Curran Associates, Inc.}, 2020, pp. 12\,449--12\,460.

\bibitem[Baevski et~al.(2022)Baevski, Hsu, Xu, Babu, Gu, and
  Auli]{baevskiData2vecGeneralFramework2022}
A.~Baevski, W.-N. Hsu, Q.~Xu, A.~Babu, J.~Gu, and M.~Auli, ``data2vec: {{A}}
  general framework for self-supervised learning in speech, vision and
  language,'' in \emph{Proceedings of the 39th {{International Conference}} on
  {{Machine Learning}}}, ser. Proceedings of machine learning research,
  K.~Chaudhuri, S.~Jegelka, L.~Song, C.~Szepesvari, G.~Niu, and S.~Sabato,
  Eds., vol. 162.\hskip 1em plus 0.5em minus 0.4em\relax {PMLR}, Jul. 2022, pp.
  1298--1312.

\bibitem[Chen et~al.(2022)Chen, Wang, Chen, Wu, Liu, Chen, Li, Kanda, Yoshioka,
  Xiao, Wu, Zhou, Ren, Qian, Qian, Wu, Zeng, Yu, and
  Wei]{chenWavLMLargescaleSelfsupervised2022}
S.~Chen, C.~Wang, Z.~Chen, Y.~Wu, S.~Liu, Z.~Chen, J.~Li, N.~Kanda,
  T.~Yoshioka, X.~Xiao, J.~Wu, L.~Zhou, S.~Ren, Y.~Qian, Y.~Qian, J.~Wu,
  M.~Zeng, X.~Yu, and F.~Wei, ``{{WavLM}}: {{Large-scale}} self-supervised
  pre-training for full stack speech processing,'' \emph{IEEE Journal of
  Selected Topics in Signal Processing}, vol.~16, no.~6, pp. 1505--1518, 2022.

\bibitem[Zhang et~al.(2023)Zhang, Han, Qin, Wang, Bapna, Chen, Chen, Li,
  Axelrod, Wang, Meng, Hu, Rosenberg, Prabhavalkar, Park, Haghani, Riesa,
  Perng, Soltau, Strohman, Ramabhadran, Sainath, Moreno, Chiu, Schalkwyk,
  Beaufays, and Wu]{zhangGoogleUSMScaling2023}
Y.~Zhang, W.~Han, J.~Qin, Y.~Wang, A.~Bapna, Z.~Chen, N.~Chen, B.~Li,
  V.~Axelrod, G.~Wang, Z.~Meng, K.~Hu, A.~Rosenberg, R.~Prabhavalkar, D.~S.
  Park, P.~Haghani, J.~Riesa, G.~Perng, H.~Soltau, T.~Strohman, B.~Ramabhadran,
  T.~Sainath, P.~Moreno, C.-C. Chiu, J.~Schalkwyk, F.~Beaufays, and Y.~Wu,
  ``Google {{USM}}: {{Scaling}} automatic speech recognition beyond 100
  languages,'' 2023.

\bibitem[Dong et~al.(2019)Dong, Wang, and
  Xu]{dongSelfattentionAlignerLatencycontrol2019}
L.~Dong, F.~Wang, and B.~Xu, ``Self-attention aligner: {{A}} latency-control
  end-to-end model for {{ASR}} using self-attention network and
  chunk-hopping,'' in \emph{2019 {{IEEE International Conference}} on
  {{Acoustics}}, {{Speech}} and {{Signal Processing}}}, 2019, pp. 5656--5660.

\bibitem[Zhang et~al.(2020)Zhang, Lu, Sak, Tripathi, McDermott, Koo, and
  Kumar]{zhangTransformerTransducerStreamable2020}
Q.~Zhang, H.~Lu, H.~Sak, A.~Tripathi, E.~McDermott, S.~Koo, and S.~Kumar,
  ``Transformer transducer: {{A}} streamable speech recognition model with
  transformer encoders and {{RNN-T}} loss,'' in \emph{2020 {{IEEE International
  Conference}} on {{Acoustics}}, {{Speech}} and {{Signal Processing}}}, 2020,
  pp. 7829--7833.

\bibitem[Fu et~al.(2023)Fu, Li, Li, Deng, Li, Fan, Chen, and
  He]{fuUFO2UnifiedPretraining2023}
L.~Fu, S.~Li, Q.~Li, L.~Deng, F.~Li, L.~Fan, M.~Chen, and X.~He, ``{{UFO2}}:
  {{A}} unified pre-training framework for online and offline speech
  recognition,'' in \emph{2023 {{IEEE International Conference}} on
  {{Acoustics}}, {{Speech}} and {{Signal Processing}}}, 2023, pp. 1--5.

\bibitem[Swietojanski et~al.(2023)Swietojanski, Braun, Can, Da~Silva, Ghoshal,
  Hori, Hsiao, Mason, McDermott, Silovsky, Travadi, and
  Zhuang]{swietojanskiVariableAttentionMasking2023}
P.~Swietojanski, S.~Braun, D.~Can, T.~F. Da~Silva, A.~Ghoshal, T.~Hori,
  R.~Hsiao, H.~Mason, E.~McDermott, H.~Silovsky, R.~Travadi, and X.~Zhuang,
  ``Variable attention masking for configurable transformer transducer speech
  recognition,'' in \emph{2023 {{IEEE International Conference}} on
  {{Acoustics}}, {{Speech}} and {{Signal Processing}}}, 2023, pp. 1--5.

\bibitem[Assmann et~al.(1982)Assmann, Nearey, and
  Hogan]{assmannVowelIdentificationOrthographic1982}
P.~F. Assmann, T.~M. Nearey, and J.~T. Hogan, ``Vowel identification:
  {{Orthographic}}, perceptual, and acoustic aspects,'' \emph{The Journal of
  the Acoustical Society of America}, vol.~71, no.~4, pp. 975--989, Apr. 1982.

\bibitem[Toscano and Allen(2014)]{toscanoWithinconsonantErrorsIsolated2014}
J.~C. Toscano and J.~B. Allen, ``Across- and within-consonant errors for
  isolated syllables in noise,'' \emph{Journal of Speech, Language, and Hearing
  Research}, vol.~57, no.~6, pp. 2293--2307, Dec. 2014.

\bibitem[Mohamed et~al.(2022)Mohamed, Lee, Borgholt, Havtorn, Edin, Igel,
  Kirchhoff, Li, Livescu, Maal{\o}e, Sainath, and
  Watanabe]{mohamedSelfsupervisedSpeechRepresentation2022}
A.~Mohamed, H.-y. Lee, L.~Borgholt, J.~D. Havtorn, J.~Edin, C.~Igel,
  K.~Kirchhoff, S.-W. Li, K.~Livescu, L.~Maal{\o}e, T.~N. Sainath, and
  S.~Watanabe, ``Self-supervised speech representation learning: {{A}}
  review,'' \emph{IEEE Journal of Selected Topics in Signal Processing},
  vol.~16, no.~6, pp. 1179--1210, 2022.

\bibitem[Borgholt et~al.(2022)Borgholt, Havtorn, Edin, Maal{\o}e, and
  Igel]{borgholtBriefOverviewUnsupervised2022}
L.~Borgholt, J.~D. Havtorn, J.~Edin, L.~Maal{\o}e, and C.~Igel, ``A brief
  overview of unsupervised neural speech representation learning,'' in
  \emph{Proceedings of 2nd {{Workshop}} on {{Self-Supervised Learning}} for
  {{Audio}} and {{Speech Processing}}}.\hskip 1em plus 0.5em minus 0.4em\relax
  {Association for the Advancement of Artificial Intelligence}, 2022.

\bibitem[Yang et~al.(2021)Yang, Chi, Chuang, Lai, Lakhotia, Lin, Liu, Shi,
  Chang, Lin, Huang, Tseng, Lee, Liu, Huang, Dong, Li, Watanabe, Mohamed, and
  Lee]{yangSUPERBSpeechProcessing2021}
S.-w. Yang, P.-H. Chi, Y.-S. Chuang, C.-I.~J. Lai, K.~Lakhotia, Y.~Y. Lin,
  A.~T. Liu, J.~Shi, X.~Chang, G.-T. Lin, T.-H. Huang, W.-C. Tseng, K.-t. Lee,
  D.-R. Liu, Z.~Huang, S.~Dong, S.-W. Li, S.~Watanabe, A.~Mohamed, and H.-y.
  Lee, ``{{SUPERB}}: {{Speech}} processing universal {{PERformance}}
  benchmark,'' in \emph{Proc. {{Interspeech}} 2021}, 2021, pp. 1194--1198.

\bibitem[Schatz et~al.(2013)Schatz, Peddinti, Bach, Jansen, Hermansky, and
  Dupoux]{schatzEvaluatingSpeechFeatures2013}
T.~Schatz, V.~Peddinti, F.~Bach, A.~Jansen, H.~Hermansky, and E.~Dupoux,
  ``Evaluating speech features with the minimal-pair {{ABX}} task: analysis of
  the classical {{MFC}}/{{PLP}} pipeline,'' in \emph{Proc. {{Interspeech}}
  2013}, 2013, pp. 1781--1785.

\bibitem[Hallap et~al.(2023)Hallap, Dupoux, and
  Dunbar]{hallapEvaluatingContextinvarianceUnsupervised2023}
M.~Hallap, E.~Dupoux, and E.~Dunbar, ``Evaluating context-invariance in
  unsupervised speech representations,'' in \emph{Proc. {{Interspeech}} 2023},
  2023, pp. 2973--2977.

\bibitem[Boothroyd and
  Nittrouer(1988)]{boothroydMathematicalTreatmentContext1988}
A.~Boothroyd and S.~Nittrouer, ``Mathematical treatment of context effects in
  phoneme and word recognition,'' \emph{The Journal of the Acoustical Society
  of America}, vol.~84, no.~1, pp. 101--114, Jul. 1988.

\bibitem[Nguyen et~al.(2020)Nguyen, {de Seyssel}, Roz{\'e}, Rivi{\`e}re,
  Kharitonov, Baevski, Dunbar, and Dupoux]{nguyenZeroResourceSpeech2020}
T.~A. Nguyen, M.~{de Seyssel}, P.~Roz{\'e}, M.~Rivi{\`e}re, E.~Kharitonov,
  A.~Baevski, E.~Dunbar, and E.~Dupoux, ``The zero resource speech benchmark
  2021: {{Metrics}} and baselines for unsupervised spoken language modeling,''
  2020.

\bibitem[{van den Oord} et~al.(2018){van den Oord}, Li, and
  Vinyals]{vandenoordRepresentationLearningContrastive2018}
A.~{van den Oord}, Y.~Li, and O.~Vinyals, ``Representation learning with
  contrastive predictive coding,'' 2018.

\bibitem[Peddinti et~al.(2015)Peddinti, Chen, Manohar, Ko, Povey, and
  Khudanpur]{peddintiJHUASpIRESystem2015}
V.~Peddinti, G.~Chen, V.~Manohar, T.~Ko, D.~Povey, and S.~Khudanpur, ``{{JHU
  ASpIRE}} system: {{Robust LVCSR}} with {{TDNNS}}, {{iVector}} adaptation and
  {{RNN-LMS}},'' in \emph{2015 {{IEEE Workshop}} on {{Automatic Speech
  Recognition}} \& {{Understanding}}}, 2015, pp. 539--546.

\bibitem[{Abdel-Hamid} et~al.(2014){Abdel-Hamid}, Mohamed, Jiang, Deng, Penn,
  and Yu]{abdel-hamidConvolutionalNeuralNetworks2014}
O.~{Abdel-Hamid}, A.-r. Mohamed, H.~Jiang, L.~Deng, G.~Penn, and D.~Yu,
  ``Convolutional neural networks for speech recognition,'' \emph{IEEE/ACM
  Transactions on Audio, Speech, and Language Processing}, vol.~22, no.~10, pp.
  1533--1545, Oct. 2014.

\bibitem[Schneider et~al.(2019)Schneider, Baevski, Collobert, and
  Auli]{schneiderWav2vecUnsupervisedPretraining2019}
S.~Schneider, A.~Baevski, R.~Collobert, and M.~Auli, ``wav2vec:
  {{Unsupervised}} pre-training for speech recognition,'' in \emph{Proc.
  {{Interspeech}} 2019}, 2019, pp. 3465--3469.

\bibitem[Bahdanau et~al.(2015)Bahdanau, Cho, and
  Bengio]{bahdanauNeuralMachineTranslation2015}
D.~Bahdanau, K.~Cho, and Y.~Bengio, ``Neural machine translation by jointly
  learning to align and translate,'' in \emph{3rd {{International Conference}}
  on {{Learning Representations}}}, {San Diego, USA}, 2015.

\bibitem[Raffel et~al.(2017)Raffel, Luong, Liu, Weiss, and
  Eck]{raffelOnlineLineartimeAttention2017}
C.~Raffel, M.-T. Luong, P.~J. Liu, R.~J. Weiss, and D.~Eck, ``Online and
  linear-time attention by enforcing monotonic alignments,'' in
  \emph{International {{Conference}} on {{Machine Learning}}}, ser. Proceedings
  of {{Machine Learning Research}}, D.~Precup and Y.~W. Teh, Eds.,
  vol.~70.\hskip 1em plus 0.5em minus 0.4em\relax {Sydney, Australia}: {PMLR},
  Aug. 2017, pp. 2837--2846.

\bibitem[Tjandra et~al.(2017)Tjandra, Sakti, and
  Nakamura]{tjandraLocalMonotonicAttention2017}
A.~Tjandra, S.~Sakti, and S.~Nakamura, ``Local monotonic attention mechanism
  for end-to-end speech and language processing,'' in \emph{Proceedings of the
  {{Eighth International Joint Conference}} on {{Natural Language
  Processing}}}, vol.~1.\hskip 1em plus 0.5em minus 0.4em\relax {Taipei,
  Taiwan}: {Asian Federation of Natural Language Processing}, Nov. 2017, pp.
  431--440.

\bibitem[Hou et~al.(2017)Hou, Zhang, and Dai]{houGaussianPredictionBased2017}
J.~Hou, S.~Zhang, and L.-R. Dai, ``Gaussian prediction based attention for
  online end-to-end speech recognition,'' in \emph{Proc. {{Interspeech}} 2017},
  2017, pp. 3692--3696.

\bibitem[Chiu and Raffel(2018)]{chiuMonotonicChunkwiseAttention2018}
C.-C. Chiu and C.~Raffel, ``Monotonic chunkwise attention,'' in \emph{6th
  {{International Conference}} on {{Learning Representations}}}, {Vancouver,
  Canada}, 2018.

\bibitem[Arivazhagan et~al.(2019)Arivazhagan, Cherry, Macherey, Chiu, Yavuz,
  Pang, Li, and Raffel]{arivazhaganMonotonicInfiniteLookback2019}
N.~Arivazhagan, C.~Cherry, W.~Macherey, C.-C. Chiu, S.~Yavuz, R.~Pang, W.~Li,
  and C.~Raffel, ``Monotonic infinite lookback attention for simultaneous
  machine translation,'' in \emph{Proceedings of the 57th {{Annual Meeting}} of
  the {{Association}} for {{Computational Linguistics}}}.\hskip 1em plus 0.5em
  minus 0.4em\relax {Florence, Italy}: {Association for Computational
  Linguistics}, Jul. 2019, pp. 1313--1323.

\bibitem[Prabhavalkar et~al.(2017)Prabhavalkar, Rao, Sainath, Li, Johnson, and
  Jaitly]{prabhavalkarComparisonSequencetosequenceModels2017}
R.~Prabhavalkar, K.~Rao, T.~N. Sainath, B.~Li, L.~Johnson, and N.~Jaitly, ``A
  comparison of sequence-to-sequence models for speech recognition,'' in
  \emph{Proc. {{Interspeech}} 2017}, 2017, pp. 939--943.

\bibitem[Chorowski et~al.(2015)Chorowski, Bahdanau, Serdyuk, Cho, and
  Bengio]{chorowskiAttentionbasedModelsSpeech2015}
J.~K. Chorowski, D.~Bahdanau, D.~Serdyuk, K.~Cho, and Y.~Bengio,
  ``Attention-based models for speech recognition,'' in \emph{Advances in
  {{Neural Information Processing Systems}} 28}, C.~Cortes, N.~D. Lawrence,
  D.~D. Lee, M.~Sugiyama, and R.~Garnett, Eds.\hskip 1em plus 0.5em minus
  0.4em\relax {Curran Associates, Inc.}, 2015, pp. 577--585.

\bibitem[Chiu et~al.(2019)Chiu, Han, Zhang, Pang, Kishchenko, Nguyen,
  Narayanan, Liao, {Shuyuan Zhang}, Kannan, Prabhavalkar, Chen, Sainath, and
  Wu]{chiuComparisonEndtoendModels2019}
C.-C. Chiu, W.~Han, Y.~Zhang, R.~Pang, S.~Kishchenko, P.~Nguyen, A.~Narayanan,
  H.~Liao, {Shuyuan Zhang}, A.~Kannan, R.~Prabhavalkar, Z.~Chen, T.~Sainath,
  and Y.~Wu, ``A comparison of end-to-end models for long-form speech
  recognition,'' in \emph{2019 {{IEEE Automatic Speech Recognition}} and
  {{Understanding Workshop}}}, 2019, pp. 889--896.

\bibitem[Panayotov et~al.(2015)Panayotov, Chen, Povey, and
  Khudanpur]{panayotovLibrispeechASRCorpus2015}
V.~Panayotov, G.~Chen, D.~Povey, and S.~Khudanpur, ``Librispeech: {{An ASR}}
  corpus based on public domain audio books,'' in \emph{2015 {{IEEE
  International Conference}} on {{Acoustics}}, {{Speech}} and {{Signal
  Processing}}}, Apr. 2015, pp. 5206--5210.

\bibitem[Tian et~al.(2020)Tian, Sun, Poole, Krishnan, Schmid, and
  Isola]{tianWhatMakesGood2020}
Y.~Tian, C.~Sun, B.~Poole, D.~Krishnan, C.~Schmid, and P.~Isola, ``What makes
  for good views for contrastive learning?'' in \emph{Advances in {{Neural
  Information Processing Systems}}}, H.~Larochelle, M.~Ranzato, R.~Hadsell,
  M.~Balcan, and H.~Lin, Eds., vol.~33.\hskip 1em plus 0.5em minus 0.4em\relax
  {Curran Associates, Inc.}, 2020, pp. 6827--6839.

\bibitem[Vaswani et~al.(2017)Vaswani, Shazeer, Parmar, Uszkoreit, Jones, Gomez,
  Kaiser, and Polosukhin]{vaswaniAttentionAllYou2017}
A.~Vaswani, N.~Shazeer, N.~Parmar, J.~Uszkoreit, L.~Jones, A.~N. Gomez,
  {\L}.~Kaiser, and I.~Polosukhin, ``Attention is all you need,'' in
  \emph{Advances in {{Neural Information Processing Systems}} 30}, I.~Guyon,
  U.~V. Luxburg, S.~Bengio, H.~Wallach, R.~Fergus, S.~Vishwanathan, and
  R.~Garnett, Eds.\hskip 1em plus 0.5em minus 0.4em\relax {Curran Associates,
  Inc.}, 2017, pp. 5998--6008.

\bibitem[Versteegh et~al.(2016)Versteegh, Anguera, Jansen, and
  Dupoux]{versteeghZeroResourceSpeech2016}
M.~Versteegh, X.~Anguera, A.~Jansen, and E.~Dupoux, ``The zero resource speech
  challenge 2015: {{Proposed}} approaches and results,'' \emph{Procedia
  Computer Science}, vol.~81, pp. 67--72, 2016.

\bibitem[Dunbar et~al.(2022)Dunbar, Hamilakis, and
  Dupoux]{dunbarSelfsupervisedLanguageLearning2022}
E.~Dunbar, N.~Hamilakis, and E.~Dupoux, ``Self-supervised language learning
  from raw audio: {{Lessons}} from the zero resource speech challenge,''
  \emph{IEEE Journal of Selected Topics in Signal Processing}, vol.~16, no.~6,
  pp. 1211--1226, 2022.

\bibitem[Hochreiter and Schmidhuber(1997)]{hochreiterLongShorttermMemory1997}
S.~Hochreiter and J.~Schmidhuber, ``Long short-term memory,'' \emph{Neural
  Computation}, vol.~9, no.~8, pp. 1735--1780, Nov. 1997.

\bibitem[He et~al.(2016)He, Zhang, Ren, and Sun]{heDeepResidualLearning2016}
K.~He, X.~Zhang, S.~Ren, and J.~Sun, ``Deep residual learning for image
  recognition,'' in \emph{2016 {{IEEE Conference}} on {{Computer Vision}} and
  {{Pattern Recognition}}}, Jun. 2016, pp. 770--778.

\bibitem[Ba et~al.(2016)Ba, Kiros, and Hinton]{baLayerNormalization2016}
J.~L. Ba, J.~R. Kiros, and G.~E. Hinton, ``Layer normalization,'' 2016.

\bibitem[Rivi{\`e}re et~al.(2020)Rivi{\`e}re, Joulin, Mazar{\'e}, and
  Dupoux]{riviereUnsupervisedPretrainingTransfers2020}
M.~Rivi{\`e}re, A.~Joulin, P.-E. Mazar{\'e}, and E.~Dupoux, ``Unsupervised
  pretraining transfers well across languages,'' in \emph{2020 {{IEEE
  International Conference}} on {{Acoustics}}, {{Speech}}, and {{Signal
  Processing}}}, 2020, pp. 7414--7418.

\bibitem[Hornik et~al.(1989)Hornik, Stinchcombe, and
  White]{hornikMultilayerFeedforwardNetworks1989}
K.~Hornik, M.~Stinchcombe, and H.~White, ``Multilayer feedforward networks are
  universal approximators,'' \emph{Neural Networks}, vol.~2, no.~5, pp.
  359--366, 1989.

\bibitem[Mallat(2016)]{mallatUnderstandingDeepConvolutional2016}
S.~Mallat, ``Understanding deep convolutional networks,'' \emph{Philosophical
  Transactions of the Royal Society A: Mathematical, Physical and Engineering
  Sciences}, vol. 374, no. 2065, Mar. 2016.

\bibitem[Shain and Elsner(2020)]{shainAcquiringLanguageSpeech2020}
C.~Shain and M.~Elsner, ``Acquiring language from speech by learning to
  remember and predict,'' in \emph{Proceedings of the 24th {{Conference}} on
  {{Computational Natural Language Learning}}}.\hskip 1em plus 0.5em minus
  0.4em\relax {Online}: {Association for Computational Linguistics}, Nov. 2020,
  pp. 195--214.

\bibitem[Chiu et~al.(2022)Chiu, Qin, Zhang, Yu, and
  Wu]{chiuSelfsupervisedLearningRandomprojection2022}
C.-C. Chiu, J.~Qin, Y.~Zhang, J.~Yu, and Y.~Wu, ``Self-supervised learning with
  random-projection quantizer for speech recognition,'' in \emph{Proceedings of
  the 39th {{International Conference}} on {{Machine Learning}}}, ser.
  Proceedings of {{Machine Learning Research}}, K.~Chaudhuri, S.~Jegelka,
  L.~Song, C.~Szepesvari, G.~Niu, and S.~Sabato, Eds., vol. 162.\hskip 1em plus
  0.5em minus 0.4em\relax {PMLR}, Jul. 2022, pp. 3915--3924.

\bibitem[Graves et~al.(2006)Graves, Fern{\'a}ndez, Gomez, and
  Schmidhuber]{gravesConnectionistTemporalClassification2006}
A.~Graves, S.~Fern{\'a}ndez, F.~Gomez, and J.~Schmidhuber, ``Connectionist
  {{Temporal Classification}}: {{Labelling}} unsegmented sequence data with
  recurrent neural networks,'' in \emph{Proceedings of the 23rd {{International
  Conference}} on {{Machine Learning}}}.\hskip 1em plus 0.5em minus 0.4em\relax
  {New York, NY, USA}: {ACM}, 2006, pp. 369--376.

\bibitem[Park et~al.(2019)Park, Chan, Zhang, Chiu, Zoph, Cubuk, and
  Le]{parkSpecAugmentSimpleData2019}
D.~S. Park, W.~Chan, Y.~Zhang, C.-C. Chiu, B.~Zoph, E.~D. Cubuk, and Q.~V. Le,
  ``{{SpecAugment}}: {{A}} simple data augmentation method for automatic speech
  recognition,'' in \emph{Proc. {{Interspeech}} 2019}, 2019, pp. 2613--2617.

\bibitem[{Kensho Technologies}()]{kenshotechnologiesPyctcdecode}
{Kensho Technologies}, ``pyctcdecode.''

\bibitem[Heafield(2011)]{heafieldKenLMFasterSmaller2011}
K.~Heafield, ``{{KenLM}}: faster and smaller language model queries,'' in
  \emph{Proceedings of the {{Sixth Workshop}} on {{Statistical Machine
  Translation}}}.\hskip 1em plus 0.5em minus 0.4em\relax {Edinburgh, Scotland}:
  {Association for Computational Linguistics}, 2011, pp. 187--197.

\bibitem[Vallat(2018)]{vallatPingouinStatisticsPython2018}
R.~Vallat, ``Pingouin: {{Statistics}} in python,'' \emph{Journal of Open Source
  Software}, vol.~3, no.~31, p. 1026, 2018.

\bibitem[Mohamed et~al.(2012)Mohamed, Dahl, and
  Hinton]{mohamedAcousticModelingUsing2012}
A.-r. Mohamed, G.~E. Dahl, and G.~Hinton, ``Acoustic modeling using deep belief
  networks,'' \emph{IEEE Transactions on Audio, Speech, and Language
  Processing}, vol.~20, no.~1, pp. 14--22, Jan. 2012.

\bibitem[Mallat(2008)]{mallatWaveletTourSignal2008}
S.~Mallat, \emph{A wavelet tour of signal processing: {{The}} sparse
  way}.\hskip 1em plus 0.5em minus 0.4em\relax {Elsevier Science}, 2008.

\end{thebibliography}

\begin{IEEEbiographynophoto}{Sean Robertson}
is a post-doctoral fellow at the University of Toronto, Canada, funded by the Data Sciences Institute. He holds a PhD and an MSc from the University of Toronto, and a BCS from the University of Toronto. He is also a Faculty Affiliate Researcher with the Vector Institute. His interests are in speech recognition and pre-training.
\end{IEEEbiographynophoto}

\begin{IEEEbiographynophoto}{Ewan Dunbar}
is an Assistant Professor in the French Linguistics program at the University of Toronto and an Affiliated Scientist with the Cognitive Machine Learning (CoML) team at the Ecole Normale Sup\'{e}rieure (ENS) in Paris and INRIA. He holds a PhD in Linguistics from the University of Maryland, College Park, an MA in Linguistics from the University of Toronto, and a BSc in Linguistics and Computer Science, also from the University of Toronto. His research focuses on speech perception, psycholinguistics, and speech technology. He is co-organizer of the Zero Resource Speech Challenge (2017--present).
\end{IEEEbiographynophoto}

\end{document}


\appendix

\noindent \emph{Pre-training versus downstream metrics}. We briefly explore the agreement between ABX-LS error rates and their corresponding upstream models' validation losses.

\begin{figure}
\centering
\begin{minipage}{0.45\textwidth}
    \includegraphics[width=\columnwidth]{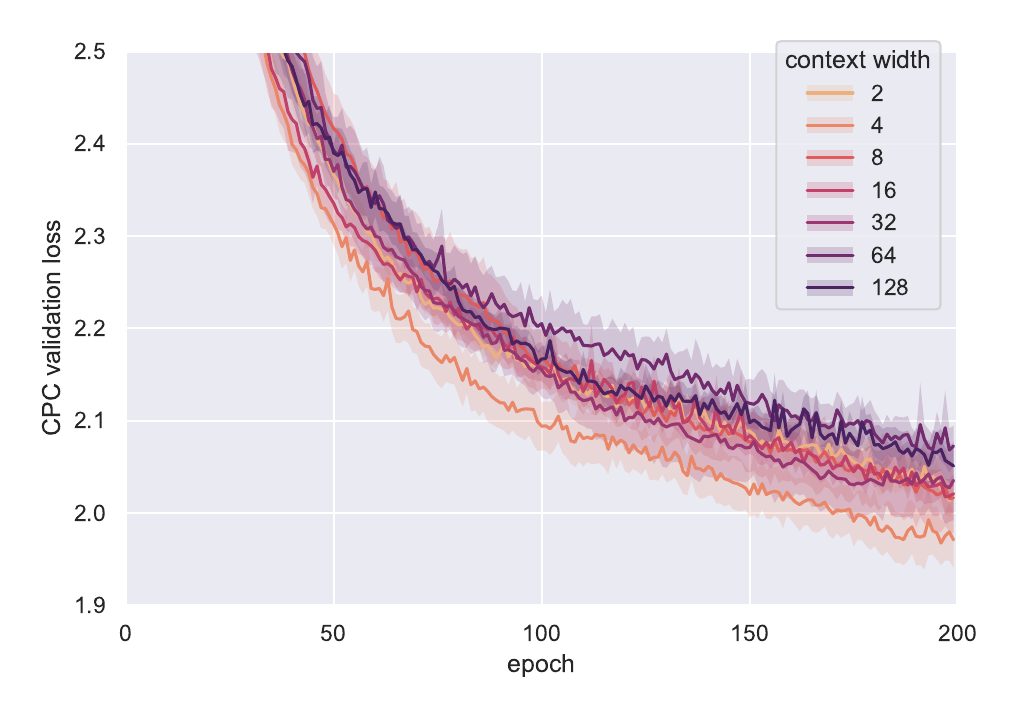}
\end{minipage}
\caption{%
    Comparing validation loss over epochs across context widths $W$ from our \emph{main} training regime, $200$ epochs. Error bars are $\pm 1$ SE over $N=5$ trained models per width.
} \label{fig:loss}
\end{figure}

\Cref{fig:loss} shows the average CPC validation loss per epoch across all the context widths from our main experimental setup.\footnote{%
    For reference, the average minimal validation losses in ascending order of context width are: $2.010$, $1.958$, $2.008$, $2.010$, $2.013$, $2.052$, and $2.043$.%
} Though the exact order of winners sometimes differ, The validation score and ABX-LS error rate agree in trend: $W = 4$ is the lowest in both, $W \geq 64$ are highest in both, and the remainder are in-between. In conjunction with our findings from our Wilcoxon signed-rank tests, we may conclude that the loss and error rate are in broad agreement.

\begin{figure}
\centering
\includegraphics[width=\columnwidth]{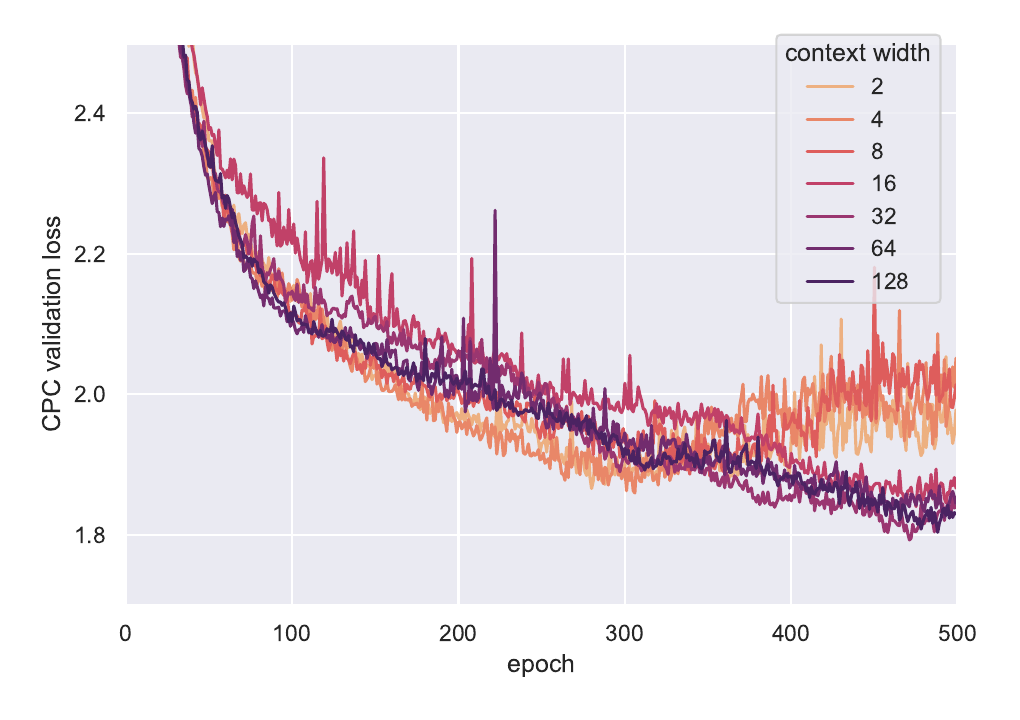}
\caption{%
    Comparing validation loss over epochs across context widths $W$ in the \emph{long train} setup. A single model per width was trained, $500$ epochs each.%
} \label{fig:loss_long}
\end{figure}

\begin{figure}
\centering
\includegraphics[width=\columnwidth]{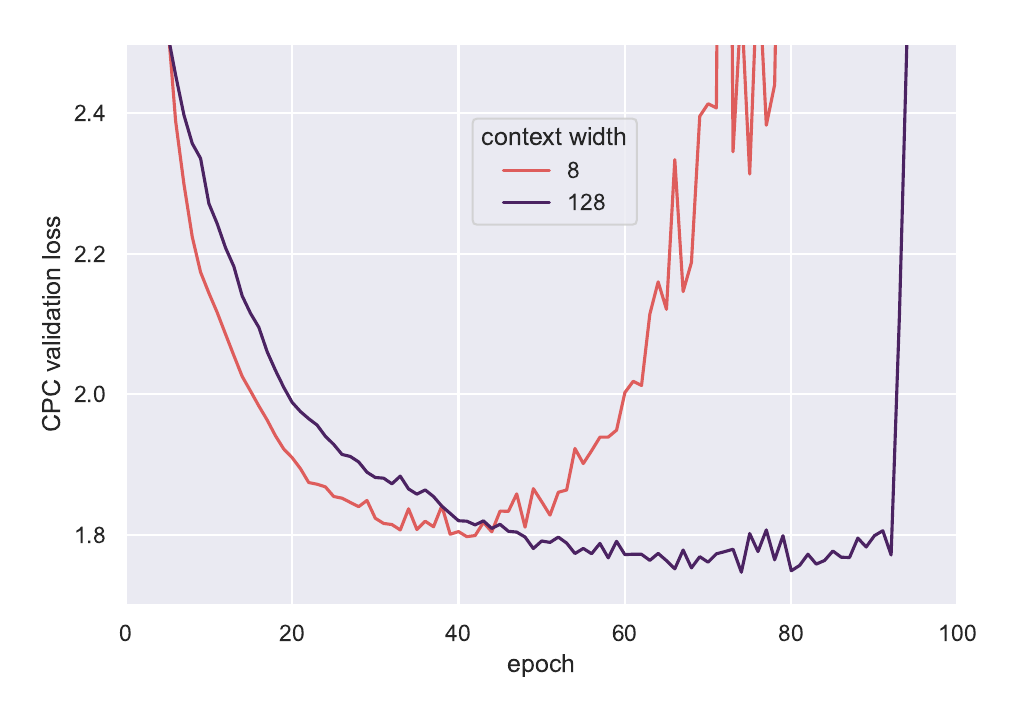}
\caption{%
    \small
    Comparing validation loss over epochs across two context widths, $W \in \{8,128\}$, in the \emph{960h} setup. One model was trained per width, $100$ epochs each.%
} \label{fig:loss_960}
\end{figure}

That agreement falters as soon as more data are added or the duration of pre-training increases. \Cref{fig:loss_long,fig:loss_960} illustrate the curves for the \emph{long train} and \emph{960h} conditions, respectively.\footnote{%
    The minima for the \emph{long train} condition are, by ascending width: $1.866$, $1.860$, $1.884$, $1.840$, $1.793$, $1.823$, and $1.804$; for the \emph{960h} condition they are: $1.798$ and $1.747$.%
} The validation loss predicts better performance from the models with longer context widths, though this is clearly not the case.

\begin{figure}
\centering
\includegraphics[width=\columnwidth]{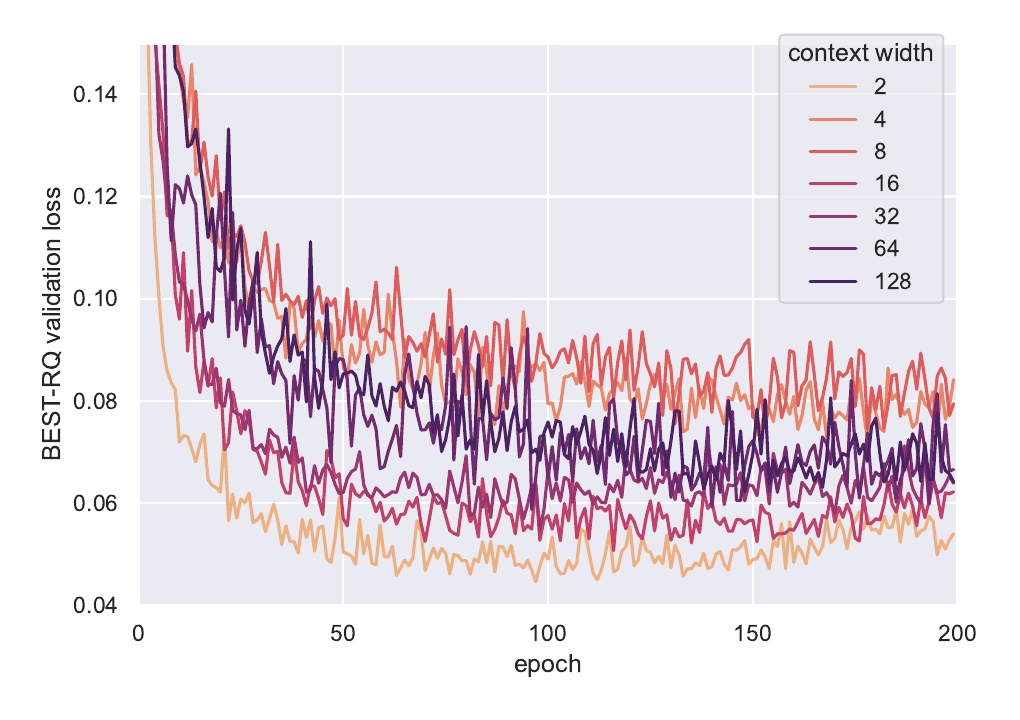}
\caption{%
    Comparing validation loss over epochs across context widths $W$ in the \emph{BEST-RQ} setup. A single model per width was trained, $200$ epochs each.%
} \label{fig:loss_bestrq}
\end{figure}

The \emph{BEST-RQ} results are similarly unhelpful: the model with the lowest loss has the second-highest error rate while the model with the lowest error rate has the second-highest validation loss.\footnote{
    Minima at $0.045$, $0.072$, $0.074$, $0.050$, $0.055$, $0.059$, $0.0622$.%
}



Overall, it appears as though there is not a strong agreement between pre-training loss and downstream performance. This could be due to a mismatch in the information each task finds important, the fact that an indeterminate amount of processing done by the predictor network is lopped off before being fed into the downstream task, or a combination of thereof. Regardless, it is clear that the pre-training loss is not a sufficient predictor of downstream performance.

\begin{figure*}
\centering
\includegraphics[width=\textwidth]{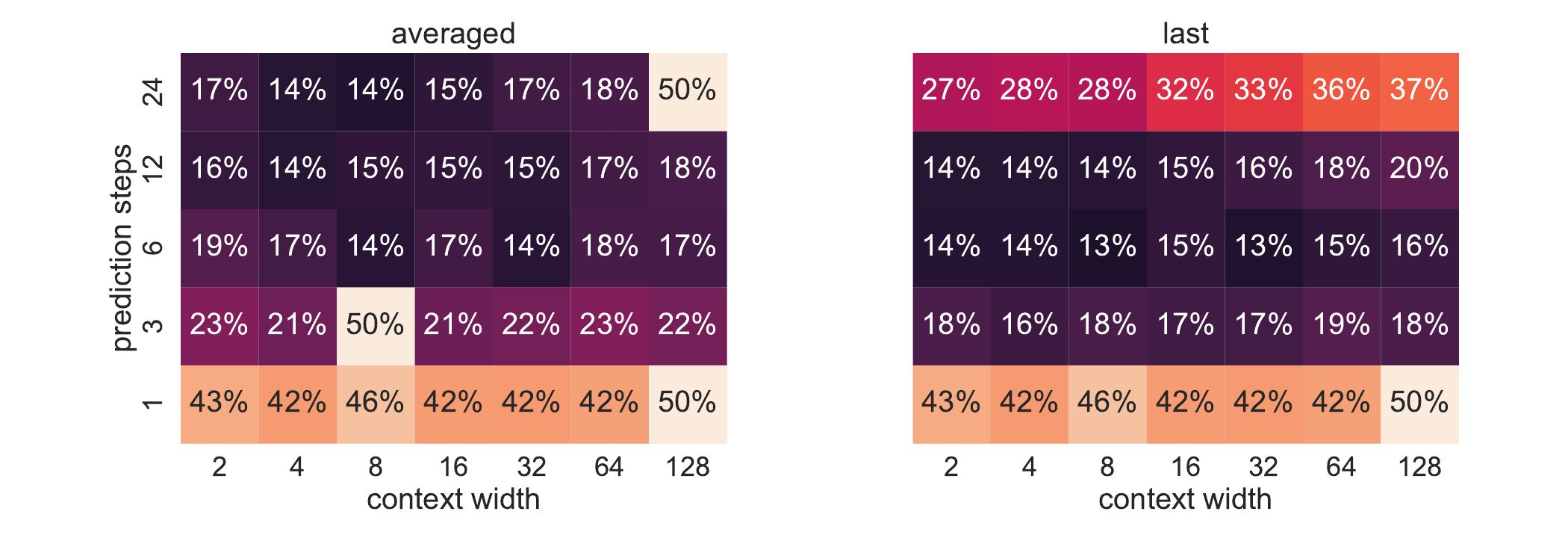}
\caption{%
    Comparing average ABX-LS performance of models of various context widths $W$ and prediction steps $S$. \emph{Left:} the classic CPC objective, with predictions over steps $1$ to $S$. \emph{Right:} CPC, but only predicting the last step $S$. Row $S=12$ on the left corresponds to the \emph{main} condition, averaged over $N=5$ models each. All other cells are just $N=1$ model each.%
} \label{fig:steps_vs_width}
\end{figure*}

\noindent \emph{Interaction effects between $S$ and $W$}. Our experiments explored the possibility of an interaction between the number of steps ahead predicted by the CPC objective, $S$, and the context width of the representations, $W$. \Cref{fig:steps_vs_width} breaks down the ABX-LS error rates by width, number of steps ahead, and whether the steps in between were predicted or not. A $50$\% error rate means the model failed to converge. In the averaged case, per-width minima cluster between $6 \leq S \leq 24$ steps. In the last-only case, those minima sit between $6 \leq S \leq 12$. Either way, there are no clear diagonal trends, suggesting first-order interactions between $S$ and $W$ are minimally impactful.